\documentclass[10pt,twocolumn,letterpaper]{article}

\usepackage[pagenumbers]{cvpr} 

\usepackage{graphicx}
\usepackage{amsmath}
\usepackage{amssymb}
\usepackage{booktabs}
\usepackage{multirow}
\usepackage{makecell}
\usepackage{amsmath,amsfonts,bm}

\def\eqref#1{equation~\ref{#1}}

\def\1{\bm{1}}

\def\rmX{{\mathbf{X}}}
\def\rmY{{\mathbf{Y}}}

\def\ve{{\bm{e}}}

\def\vq{{\bm{q}}}

\def\vx{{\bm{x}}}
\def\vy{{\bm{y}}}

\def\mA{{\bm{A}}}

\def\mW{{\bm{W}}}

\DeclareMathAlphabet{\mathsfit}{\encodingdefault}{\sfdefault}{m}{sl}
\SetMathAlphabet{\mathsfit}{bold}{\encodingdefault}{\sfdefault}{bx}{n}

\newcommand{\Ls}{\mathcal{L}}
\newcommand{\R}{\mathbb{R}}

\usepackage{algorithm}
\usepackage{algorithmic}
\usepackage[pagebackref,breaklinks,colorlinks]{hyperref}

\newcommand*{\affaddr}[1]{#1} 
\newcommand*{\affmark}[1][*]{\textsuperscript{#1}}
\newcommand*{\email}[1]{\small{\texttt{#1}}}
\usepackage[capitalize]{cleveref}
\crefname{section}{Sec.}{Secs.}
\Crefname{section}{Section}{Sections}
\Crefname{table}{Table}{Tables}
\crefname{table}{Tab.}{Tabs.}

\def\cvprPaperID{2218} 
\def\confName{CVPR}
\def\confYear{2023}

\begin{document}

\title{Model Extraction Attacks on Split Federated Learning}

\author{%
Jingtao Li\affmark[1], Adnan Siraj Rakin\affmark[1], Xing Chen\affmark[1], Li Yang\affmark[1], Zhezhi He\affmark[2], Deliang Fan\affmark[1], Chaitali Chakrabarti\affmark[1]\\
\affaddr{\affmark[1] School of Electrical Computer and Energy Engineering, Arizona State University, Tempe, AZ}\\
\affaddr{\affmark[2] Department of Computer Science and Engineering, Shanghai Jiao Tong University, Shanghai}
\\
\email{\affmark[1]\{jingtao1, asrakin, xchen382, lyang166, dfan, chaitali\}@asu.edu};\; \email{\affmark[2]\{zhezhi.he\}@sjtu.edu.cn}
}
\maketitle

\begin{abstract}
Federated Learning~(FL) is a popular collaborative learning scheme involving multiple clients and a server. FL focuses on protecting clients' data but turns out to be highly vulnerable to Intellectual Property~(IP) threats. Since FL periodically collects and distributes the model parameters, a free-rider can download the latest model and thus steal model IP. Split Federated Learning~(SFL), a recent variant of FL that supports training with resource-constrained clients, splits the model into two, giving one part of the model to clients~(client-side model), and the remaining part to the server~(server-side model). Thus SFL prevents model leakage by design. Moreover, by blocking prediction queries, it can be made resistant to advanced IP threats such as traditional Model Extraction~(ME) attacks. While SFL is better than FL in terms of providing IP protection, it is still vulnerable. In this paper, we expose the vulnerability of SFL and show how malicious clients can launch ME attacks by querying the gradient information from the server side. We propose five variants of ME attack which differs in the gradient usage as well as in the data assumptions. We show that under practical cases, the proposed ME attacks work exceptionally well for SFL. For instance, when the server-side model has five layers, our proposed ME attack can achieve over 90\% accuracy with less than 2\% accuracy degradation with VGG-11 on CIFAR-10.
\end{abstract}

\section{Introduction}
\label{sec:intro}

\begin{figure}[htbp]
\centering
\begin{subfigure}{0.9\columnwidth}
  \centering
  \includegraphics[clip,width=0.95\columnwidth]{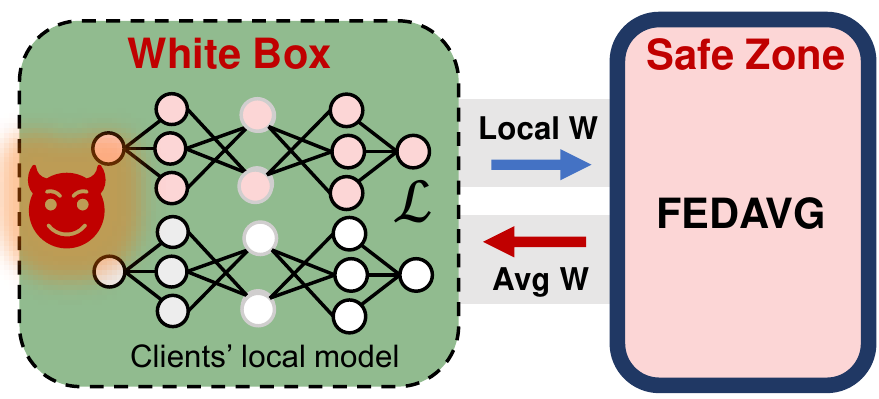}
  \label{fig:SFL_sub1}
\caption{IP threat in Federated Learning}
\end{subfigure}%
\hfill
\begin{subfigure}{0.9\columnwidth}
  \centering
  \includegraphics[clip,width=0.95\columnwidth]{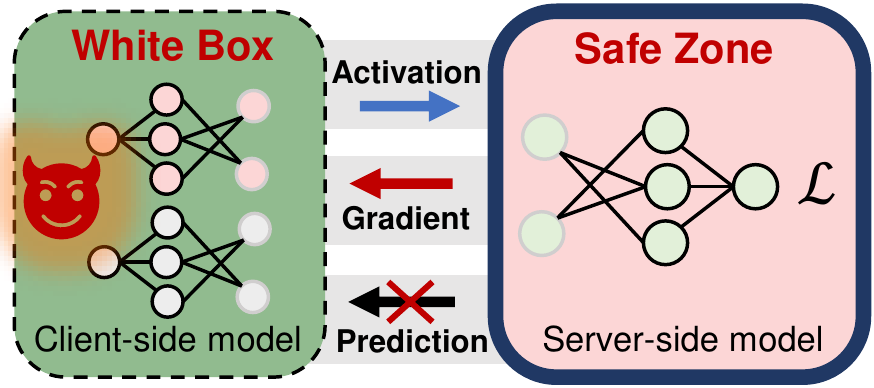}
  \label{fig:SFL_sub2}
\caption{IP threat in Split Federated Learning}
\end{subfigure}
\caption{IP threats in Federated Learning. (a) FL suffers from direct model leakage. (b) SFL prevents direct model leakage and is resistant to existing ME attacks by blocking prediction query.}
\label{fig:SFL_extraction}
\end{figure}

Federated Learning~(FL) has become increasingly popular thanks to its ability to protect users’ data and comply with General Data Protection Regulation policy. In FedAvg~\cite{mcmahan2017communication}, which is the most representative FL scheme, clients locally update their model copies, send them to the server which then aggregates the model parameters and sends the aggregated model back to the clients.
Such a setting only allows model parameters to be shared with the server, and direct data sharing is avoided.
However, we notice that FL is vulnerable to Intellectual Property~(IP) threat as a malicious client can acquire the entire model for free~(\cref{fig:SFL_extraction}~(a)). Considering the cost of hosting the central server and effort in co-ordinate the training, and the danger in using stolen model to trigger adversarial attacks~\cite{goodfellow2014explaining}, the lack of model IP protection in FL is a significant issue.

Split Federated Learning~(SFL) scheme~\cite{thapa2020splitfed} is a variant of FL for training with resource-constrained clients, where the neural network is split into a client-side model and a server-side model. Each client only computes the forward/backward propagation of the smaller client-side model while the server, which has more compute resources, computes the forward/backward propagation of the larger server-side model.
SFL follows the same model averaging routine as FL to synchronize the model. 
Unlike FL which passes on the entire model over to the clients, SFL preserves the server-side model and prevents the model from direct leakage.
Moreover, according to our investigation, SFL is resistant to existing Model Extraction~(ME) attack~\cite{tramer2016stealing, jagielski2020high} where querying a publicly accessible prediction API is needed. If 
SFL protocol does not allow prediction query access, all the prior ME attacks fail to succeed, as illustrated in~\cref{fig:SFL_extraction}~(b).


We intend to answer the following questions: \textbf{``Is SFL resistant to ME attacks? If not, then how can it be made resistant?''} To conduct the study, we test the IP threat resistance of SFL by attacking it. However, existing ME attacks~\cite{correia2018copycat, orekondy2019knockoff, truong2021data} cannot be performed since SFL presents an unique threat model where the access to prediction results is blocked.

In this paper, we propose five novel ME attacks specially designed under SFL's unique threat model. They are listed as Craft-ME, GAN-ME, GM-ME, Train-ME, and SoftTrain-ME. 
These ME attacks cover different gradient usage and data assumptions.
We also consider both train-from-scratch and fine-tuning SFL applications since ME attacks behave differently in these two cases. We benchmark the performance of five ME attacks on SFL for these two cases and find that when the number of layers in the server-side model~($N$) is small, ME attacks can succeed even without any data. However when $N$ is large, ME attacks fail badly, but SFL schemes with a large $N$ are not practical because fewer layers in client-side model could result in clients' data being compromised. 
Thus, \textbf{SFL is not inherently resistant to ME attacks}.
To answer the \textbf{second question}, we find that using L1 regularization on client-side model can improve SFL schemes' resistance to ME attacks.
In summary, we make the following contributions:

\begin{itemize}
    
    
    \item We are the first to show that SFL is not immune to ME attacks. We define a practical threat model where prediction query is blocked and attacker has white-box assumption on client-side model and gradient query access. We propose five novel ME attacks under this threat model.  We demonstrate how the white-box assumption on client-side model and gradients information can be used to extract the model. To the best of our knowledge, this is the first work that performs a comprehensive study on ME attacks in SFL.
    
    
    \item We study the effect of size of server-side model in SFL  on the success of ME, we find that when server-side model is small, our proposed ME attacks are successful. For a 5-layer-in-server VGG-11 SFL model, even without original training data, attacker can derive a surrogate model with around 85\% accuracy on CIFAR-10. 
    But when client-side model is small, we find that the ME attacks are hard to succeed. However, the small client-side model compromises clients' data and hence is not a good design choice.
    
    \item  To make SFL resistant to ME attacks, we provide a potential ME defense based on L1 regularization and show how it reduces the ME attack performance.

\end{itemize}

\section{Related Work}

\subsection{Model-split Learning Schemes}
The key idea for model-split learning schemes is to split the model so that part of it is processed in the client and the rest is offloaded to the server.
This idea was first proposed in \cite{kang2017neurosurgeon, teerapittayanon2017distributed, liu2018edgeeye} for inference tasks and extended by
\cite{gupta2018distributed} for split learning, a collaborative multi-client neural network training. However, the round-robin design in~\cite{gupta2018distributed}  need clients to learn sequentially and thus required long training time. 

\subsection{Split Federated Learning}
In SFL scheme, clients process their local models in parallel and perform periodic synchronization as in FedAvg~\cite{mcmahan2017communication}.
The detailed process of SFL is shown in Algorithm~\ref{alg:sfl}, where we reference to the SFL-V2 scheme~\cite{thapa2020splitfed}. \textbf{We define $N$ as the number of layers in server-side model, as the key design parameter}.

At the beginning of each epoch, server performs the synchronization of client-side model and sends the updated version to all clients. Then, clients perform forward propagation locally till layer $L-N$~(the last layer of client-side model), sending the intermediate activation $\mA_i$ to the server~(line 8). Server accepts the activation and label $\vy_i$ sent from clients, and uses them to calculate the loss and initiates the backward process (line 9). 
The backward process~(line 10) consists of several steps: server performs backward propagation on the loss, updates server-side model and sends back gradient $\nabla_{\mA_i} \Ls$ to clients. Clients then continue the backward propagation on their client-side model copies and perform model updates accordingly. 


\begin{algorithm}[t]
\caption{Split Federated Learning} 
\label{alg:sfl}
\begin{algorithmic}[1]
\REQUIRE $\quad$ For $M$ clients, instantiate private training data ($\rmX_i, \rmY_i$) for $1, 2, ..., M$. Server-side model $S$ has $N$ layers and client-side model $C_{i}$ has $L-N$ layers.

\STATE initialize $C_{i}, S$
\FOR{epoch $t\gets 1$ to num\_epochs}{
    \STATE $C^{*} = \frac{1}{M}\sum_{i=1}^{M}{C_{i}}$  \hfill\COMMENT{\textcolor{blue}{Model Synchronization}}
    \STATE $C_{i} \gets C^{*} $ for all $i$
    \FOR{step $s\gets 1$ to num\_batches }{
    \FOR{client $i\gets 1$ to $M$ \textbf{in Parallel}}{
    
    \STATE data batch ($\vx_i, \vy_i$) $\gets$ ($\rmX_i, \rmY_i$)
    \STATE $\mA_i = C_{i}(\mW_{C_i};\vx_i)$ \hfill\COMMENT{\textcolor{blue}{Client forward; send $\mA_i$ to Server}}
    }
    \ENDFOR
    \\\hrulefill
    \FOR{client $i\gets 1$ to $M$ \textbf{in Sequential}}{
    \STATE $\Ls = \Ls_{CE}(S(\mW_{S};\mA_i), \vy_i)$ \hfill\COMMENT{\textcolor{blue}{Server forward}}
    
    \STATE $\nabla_{\mA_i} \Ls  \gets$ back-propagation
    \hfill\COMMENT{\textcolor{blue}{Server backward, send $\nabla_{\mA_i} \Ls$ to Client}}
    }
    \STATE Update $\mW_{S}$;
    \ENDFOR
    
    \hrulefill
    \FOR{client $i\gets 1$ to $M$ \textbf{in Parallel}}{
    \STATE $\nabla_{\vx_i} \Ls  \gets$ back-propagation
    \hfill\COMMENT{\textcolor{blue}{Client backward}}
    \STATE Update $\mW_{C_i}$;
    }
    \ENDFOR

    }
\ENDFOR
}
\ENDFOR
\end{algorithmic}

\end{algorithm}

\subsection{Model Extraction Attack}
In SFL, the model is split and so IP threat due to directly downloading the model is non-existent. However, there exists advanced IP threats due to ME attacks.
Such attacks are first demonstrated in \cite{tramer2016stealing}, and the follow-up work \cite{jagielski2020high} shows that high fidelity and accurate model can be obtained with very few model prediction queries.

A successful ME attack not only breaches the model IP, but also makes the model more vulnerable to attacks. ME attack can support transferable adversarial attacks~\cite{goodfellow2014explaining}, mainly targeted ones~\cite{madry2017towards} against the victim model. A high-fidelity surrogate model also be used to perform bit-flip attacks~\cite{rakin2019bit}; for instance, a few bit flips on model parameters can degrade ResNet-18 model accuracy to below 1\%.

\subsection{Data protection in SFL}
Similar to FL, SFL scheme protects clients' data by not sending it directly to the server. However, data protection in SFL can be compromised by attacks such as MI attacks. In model-based MI attack~\cite{fredrikson2015model}, the attacker trains an inverted version of client-side model and can directly reconstruct raw inputs from the intermediate activation. Recent works~\cite{vepakomma2020nopeek, li2022ressfl} provide practical ways to mitigate MI attacks. However, they cannot achieve satisfactory mitigation when the client-side model has very few number of layers (less than 3 in a VGG-11 model).



\begin{figure*}[htbp]
\centering
\begin{subfigure}{.66\columnwidth}
  \centering
  \includegraphics[width=0.9\linewidth]{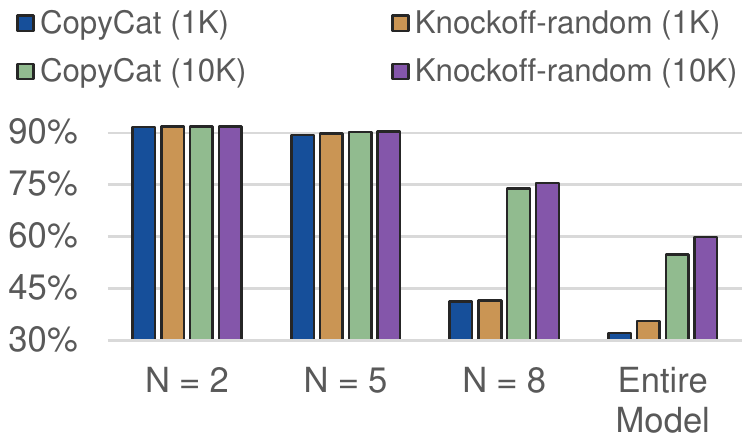}
  \label{fig:noblock_sub1}
\caption{}
\end{subfigure}%
\begin{subfigure}{.66\columnwidth}
  \centering
  \includegraphics[width=0.9\linewidth]{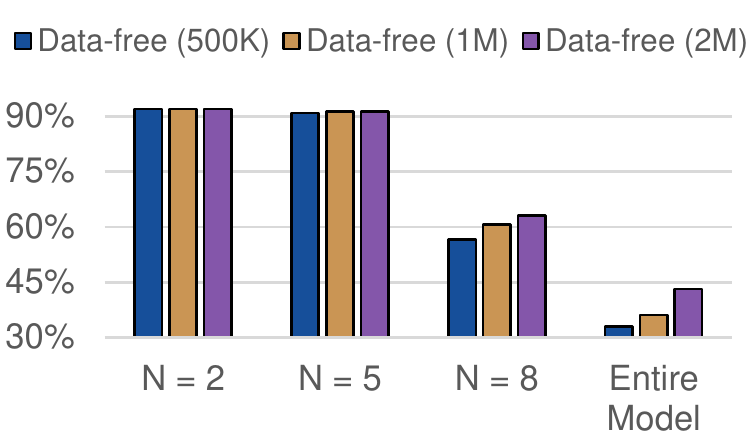}
  \label{fig:noblock_sub2}
\caption{}
\end{subfigure}
\begin{subfigure}{.66\columnwidth}
  \centering
  \includegraphics[width=0.9\linewidth]{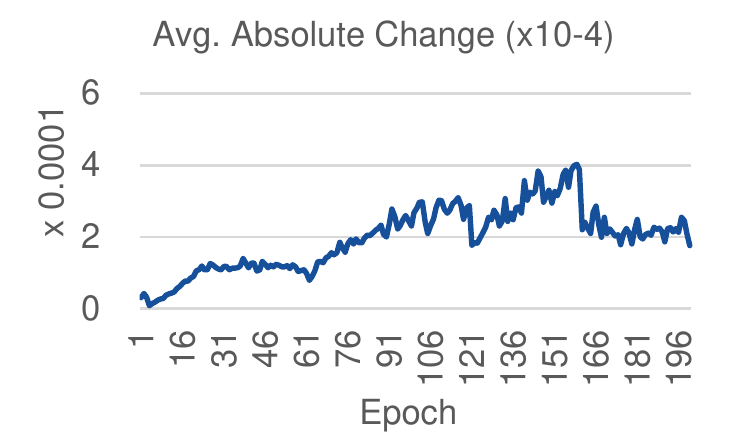}
  \label{fig:noblock_sub3}
\caption{}
\end{subfigure}
\caption{Case study VGG-11 CIFAR-10 model with different $N$: If prediction API access is allowed, existing ME attacks are very successful on SFL which suggests prediction API access should be blocked. (a) ME attacks using CIFAR-100 as auxiliary dataset; (b) Data-free ME attack that demonstrates ME attack on part of the model is much easier than ME attack on the entire model; (c) Inconsistent gradient problem in training-from-scratch SFL. The y-axis denotes the change in gradient (lower means more consistent) for the same inputs in different epochs.}
\label{fig:traditional_ME}
\end{figure*}

\section{Threat Model}
\label{sec:attacks}

\subsection{Attacker Assumptions}
\textbf{Objective.} 
According to \cite{jagielski2020high}, there are three  model extraction (ME) attack objectives: i) functional equivalence, ii) high accuracy, and iii) high fidelity. Since achieving functional equivalence is difficult in practical applications most of the existing practical ME attacks focus on achieving high accuracy and fidelity. To achieve the accuracy goal, the attacker aims to obtain a model that maximizes the prediction correctness and to achieve the fidelity goal, the attacker aims to derive a model with a similar decision boundary as the victim model before launching adversarial attacks~\cite{biggio2013evasion}.

\textbf{Data Assumption.} We assume that attackers' data assumption can fall into three categories: (1) noise data, (2) natural auxiliary data or (3) limited amount of training data.
Having noise data represents cases where the attacker uses randomly generated noise data. This case can happen when the attacker participates as a ``free-rider'' without contributing any data, or when the attacker does not have a \textit{similar enough} dataset. The second case is motivated by \cite{truong2021data} where it is shown that it is better to use random noise rather than use a drastically different dataset. In natural auxiliary data assumption, the attacker has an auxiliary dataset that is similar but with different labels from the victim's training data. For example, CIFAR-100 is such an auxiliary dataset for CIFAR-10. Furthermore, we assume a practical case where the attacker has only a subset of original training data.

\textbf{Capabilities.} We assume the attacker participate in a multi-client SFL scheme 
as outlined in~\cref{fig:SFL_extraction}. We assume the entire model has a total of $L$ layers (or layer-like blocks, i.e. BasicBlock in ResNet) out of which the server processes $N$ layers.
The attacker holds \textit{white-box assumption} on the client-side model~(consists of $L-N$ layers), that is, it knows the exact model architecture and parameters for those layers. The attacker holds a \textit{grey-box assumption} on the $N$-layer server-side model, that is, it knows its architecture and loss function
while the model parameters are unknown. 
Also, we assume \textit{server blocks the prediction queries} thus neither logits nor prediction labels are accessible by clients during training, but server \textit{allows gradient queries} to let client-side models be updated. 
Based on a client's activation $\mA=C(\vx)$ and its label $\vy$, gradient information $\nabla_\mA \Ls$ is computed and sent back to clients.

\subsection{Analysis}


\textbf{Reasons to block prediction APIs.} 
Allowing predictions APIs makes SFL vulnerable. This is particularly so since according to the white box assumption, SFL gives away the client-side model and so the attacker only needs to extract the server-side model to reveal the entire model. This results in an easier problem setting than most traditional ME attacks' assumption.
Under this easier problem setting, existing ME attacks can be very successful if  prediction APIs are \textbf{not} blocked.
Specifically, we investigate CopyCat CNN~\cite{correia2018copycat}, Knockoff-random~\cite{orekondy2019knockoff} and data-free ME~\cite{truong2021data}.
As shown in~\cref{fig:traditional_ME}~(a), with auxiliary data (CIFAR-100) and enough query budget, both attacks derive a surrogate model with very high accuracy even for a large $N$ setting.
Moreover, attacker with noise data can also succeed with data-free ME 
as shown in~\cref{fig:traditional_ME}~(a). When the query budget is equal to 2 million, the data-free ME can extract the model with high accuracy even when $N$ is equal to 5. This justifies the reason why SFL should block prediction APIs.

\textbf{Ensuring consistency of gradient query.}
We find Gradient consistency plays an important role for our proposed ME attacks.
For fine-tuning applications~\cite{park2021federated}, attackers get consistent gradient information from gradient query, as server-side model parameters are frozen or updated with a very small learning rate.
However, for a training-from-scratch usage, queries to SFL model obtain inconsistent gradient information as the server-side model drastically changes during training. As shown in~\cref{fig:traditional_ME}~(c), for the same query input, the gradient is drastically different in different epochs.

\section{Proposed Model Extraction Attack}

\begin{figure}[htbp]
\centering
\begin{subfigure}{.95\columnwidth}
  \centering
  \includegraphics[width=1.0\linewidth]{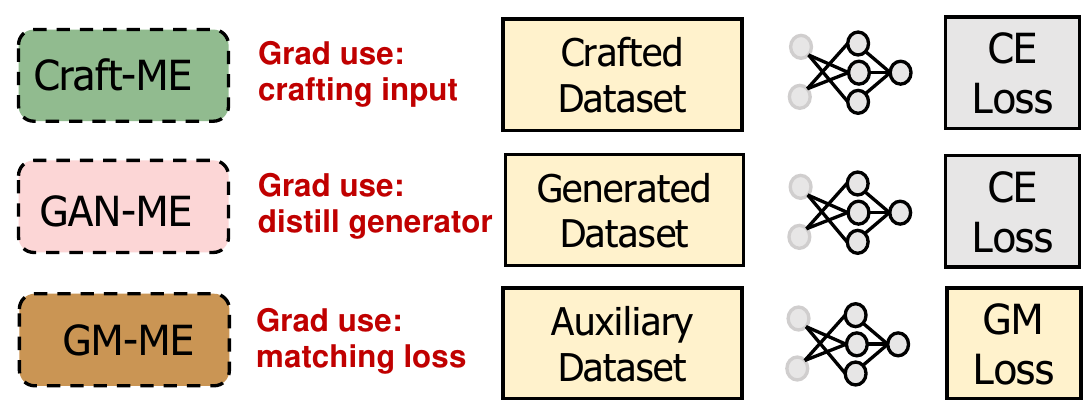}
  \label{fig:attack_sub1}
\caption{}
\end{subfigure}%
\hfill
\begin{subfigure}{.95\columnwidth}
  \centering
  \includegraphics[width=1.0\linewidth]{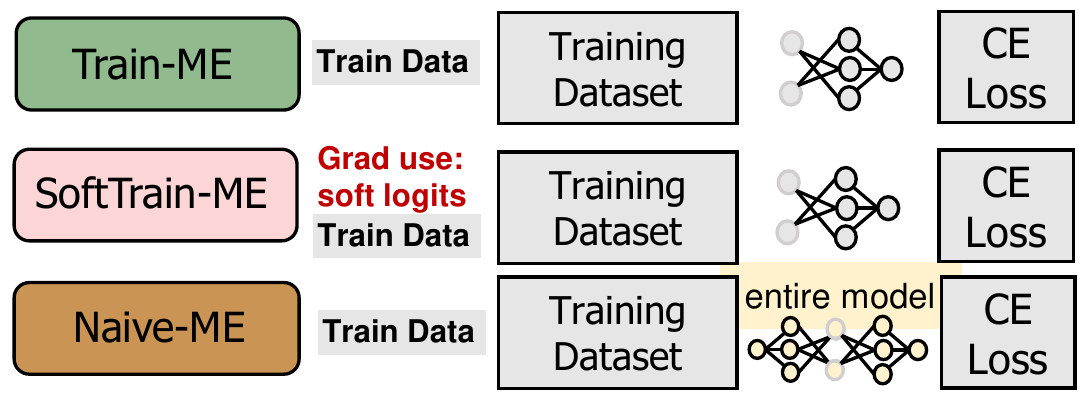}
  \label{fig:attack_sub2}
\caption{}
\end{subfigure}
\caption{Demonstration of proposed ME attacks: (a) ME attacks without training data. (b) ME attacks with training data.}
\label{fig:attacks}
\end{figure}

\begin{table*}[htbp]
\caption{Model Extraction Attack Methods in SFL}
\centering
\resizebox{0.95\linewidth}{!}{
\begin{tabular}{lcccc}
 \toprule
 \textbf{Method} & \textbf{Data Assumption} & \textbf{Prediction Query}  & \textbf{Gradient Usage} & \textbf{Client-side model Usage}\\
\midrule
Existing MEs & Varies & \textcolor{red}{Required} & None/Assistive & None\\
Naive Baseline & Limited Training & Not Required & None & None\\
\midrule
 Craft-ME & Noise Data & Not Required & Data Crafting& Initialization\\
 GAN-ME & Noise Data & Not Required & Data Generator& Initialization\\
 GM-ME & Natural Auxiliary & Not Required & Gradient Matching& Initialization\\
 \midrule
 Train-ME & Limited Training & Not Required & None & Initialization\\
 SoftTrain-ME & Limited Training & Not Required & Soft Label Crafting& Initialization\\
 
 \bottomrule
\end{tabular}
}
\label{tab:attacks}
\end{table*}

Previously, \cite{milli2019model} demonstrated that using gradients to reveal one-layer linear transformation is trivial. Given $f(\vx) = \mW^T \vx$, one can directly infer $\mW$ from a single gradient query given that $\mW^T = \nabla_\vx f(\vx)$.
However, using gradient only can go no further than one layer. \cite{milli2019model} shows that to recover a two-layer ReLU network of the form $f(\vx) = \sum_{n=1}^{h} g(\vx)_i\mW_i\mA^T_i\vx$, where $g(\vx) = \mathbb{1}$ $\{\mA\vx > 0\}$, $\mA$ is of $\R^{h\times d}$ and $\mW$ is of $\R^h$, using input gradient can recover the absolute value of normal vectors $|\mW_i\mA_i|$ for $i \in [h]$. In order to get the sign information of $\mW_i\mA_i$, prediction query is required which is not supported by  SFL's threat model.

So in this paper, we investigate approximate ME attacks that differ in the data assumptions, gradient usage and loss choices. We propose five novel ME attacks as shown in \cref{fig:attacks}. Each row corresponds to a ME attack from start to finish. For example, Craft-ME first queries the gradients to craft a dataset and uses the crafted dataset to train the surrogate server-side model from scratch using cross-entropy loss.
Despite the differences, the five proposed attack methods all train a randomly initialized surrogate server-side model from scratch. For comparison, we also include a naive baseline that ignore white-box assumption and gradients by directly training the entire model from scratch. 
We provide the detailed requirement of traditional ME attacks, naive ME attacks and our proposed five ME attacks in Table~\ref{tab:attacks}.

\subsection{ME Attacks without Training Data}



First, we proposes three attacks under weaker data assumptions when the attacker has noise data and natural auxiliary data.

\textbf{Crafting model extraction (Craft-ME).} 
Inspired by \cite{han2018co}, where data-label pairs (referred to as instances) with small-loss are shown to present useful guidance for knowledge distillation, we 
propose a simple method to craft small-loss instances using gradient queries and use them to train the surrogate model.
We initialize random input $\vx_r$ for every class label $c$, and use the gradient $\nabla_{\vx_r} \Ls$ to update $\vx_r$. For each input, updating is repeated for a number of steps.
By varying label $c$, a collection of small-loss instances is derived during SFL training. Then, a surrogate model is trained from scratch on these small-loss instances. 

\textbf{GAN-based model extraction (GAN-ME).} 
Recent work~\cite{truong2021data} proposes a GAN-based approach for data-free ME. The key idea is to use a generator $G$ to continually feed fake inputs to the victim model $V$ and surrogate model $S$, and use confidence score matching to let $S$ approach $V$. 
However, the confidence score matching needs prediction query which is not allowed in our case. Thus, we adapt the GAN-based method to gradient-query-only case and propose a two-step method: 
First, a conditional-GAN~(c-GAN) model $G(z|c)$ is initialized. The attacker trains the generator during victim model's training by generating fake data $\vx_f$ and label $c$ and performing gradient queries to update $G$. 
After the training is done, generator $G$ is used to supply small-loss instances ($\vx_f, c$) to train the surrogate model (the unknown part). We observe a serious mode collapse problem during the GAN training. So we utilize the distance-aware training introduced in \cite{yang2019diversity}, to encourage the c-GAN to generate more diverse small-loss instances.
In the new method, the training of the generator is not based on a min-max game, 
or on traditional GAN training. Instead, it simply trains the generator toward minimizing cross-entropy loss.
While the generator $G$ fails to generate natural-looking inputs even upon convergence, it generates abundant small-loss instances for every label, and divergence loss helps it generate a variety of outputs. During SFL training, the generator  adjusts  to the changing server-side model. 

\textbf{Gradient matching model extraction (GM-ME).}
Gradient matching (GM) in ME attack has been investigated in \cite{jagielski2020high, milli2019model} and is used in combination with prediction query to improve the extraction performance. Since, in SFL, prediction query is not allowed, we navigate this strict threat model's restriction by adopting gradient matching (GM) loss.
For a given label $\vy_i$, GM loss has the following form:
\begin{equation}
    \Ls_{GM} = |\nabla_{\vx_i}\Ls(S(C(\vx_i)), \vy_i) - \nabla_{\vx_i} \Ls(V(C(\vx_i)), \vy_i)|^2_2
\end{equation}
where, $\vx_i$ denote inputs, $C$ denotes client-side model, $S$ and $V$ denotes the surrogate model and victim model, respectively. For each input, attacker would query gradients with different label $\vy_i$ to get as much information as possible. 
This attack performs extremely well for small $N$ but degrades significantly for a larger $N$. Its performance also depends on the domain similarity between the auxiliary dataset and the victim dataset. 

\subsection{ME Attacks with Training Data} Next we describe the strongest data assumption case, i.e., the attacker has a subset of training data,

\textbf{Training-based model extraction (Train-ME).} For attackers with a subset of the training data, derivation of an accurate surrogate model can be done using supervised learning (through minimizing the cross entropy loss on the available data). We call this Train-ME, similar idea is also adopted in \cite{fu2022label} to extract the entire model of the other party.
Train-ME only relies on the white-box assumption of the client-side model, using it to  initialize the surrogate model and does not need to use the gradient query at all. Surprisingly, it is one of the most effective ME attacks.


\begin{table*}[htbp]
\renewcommand*{\arraystretch}{1.3}
\caption{ME attack performance on SFL on fine-tuning and training-from-scratch applications. The victim is a VGG-11 model on CIFAR-10 with 91.89\% validation accuracy. For Train, SoftTrain and Naive baseline, for the fine-tuning setting, data assumption is 1K training data (randomly sampled), and for the train-from-scratch setting, the number of clients is 10 and each client has 5K training data.}
\label{tab:stable_brief}
\begin{center}
\begin{small}
\resizebox{1.0\linewidth}{!}{
\begin{tabular}{|l|c|ccc|ccc|ccc|ccc|} 
 \hline
 \multirow{2}{*}{Metric} & \multirow{2}{*}{$N$} &\multicolumn{6}{c|}{\textbf{Fine-tuning}}  & \multicolumn{6}{c|}{\textbf{Training-from-scratch}}\\
 \cline{3-14}
 & &Craft &GAN &GM &Train &SoftTrain &Naive&Craft &GAN &GM &Train &SoftTrain&Naive\\
 \hline
 \multirow{3}{*}{\makecell{Accuracy\\(\%)}} & 2 &91.64&91.86&92.02&92.05&91.99&49.64&85.99 &85.99 &53.06 &90.58 &90.31&72.63\\
 & 5 &83.46&84.93&80.28&90.82&90.48&49.64&35.58 &40.03 &12.13 &89.86 &87.02&72.63\\
 & 8&35.48&18.82&12.45&70.28&71.32&49.64&15.34 &17.49 &10.88 &78.64 &56.78&72.63\\
 \hline
 \multirow{3}{*}{\makecell{Fidelity\\(\%)}} & 2&98.23&98.42&99.87&99.29&99.10&50.62&92.37 &89.59 &54.63 &99.34 &98.87&72.62\\
 & 5&86.32&87.49&84.33&94.84&94.67&50.62&41.32 &38.72 &11.87 &95.40 &89.83&72.62\\
 & 8&36.11&18.62&12.63&71.79&72.45&50.62&15.63 &17.44 &10.67 &80.01 &57.78&72.62\\
 \hline
\end{tabular}
}
\end{small}
\end{center}
\end{table*}

\textbf{Gradient-based soft label training model extraction (Soft-train-ME).}  If gradient query is allowed and a subset of training data is available, the attacker can achieve better ME attack performance compared to Train-ME. 
To utilize gradients, a naive idea is to combine the GM loss with cross-entropy loss in Train-ME. However, our initial investigation shows they are not compatible; the cross-entropy loss term usually dominates and the GM loss even hurts the performance.
An alternative approach is to use soft label. We build upon the method in \cite{gu2020introspective} which shows that gradient information of incorrect labels is beneficial in knowledge distillation, and use it for surrogate model training. Specifically, for each input $\vx_i$, gradients of the ground truth label as well as incorrect labels are collected $N_C$ times, where $N_C$ is the number of classes.
For an input $\vx_i$ with true label $c$, its soft label $q_i^k$ of $k$-th ($k \neq c$) label is computed as follows:\\
\begin{equation}
    q_i^k = (1-\alpha) * \frac{cos(\ve^k, \ve^c)}{ \sum_{m=1, m \neq c}^{N_C} (cos(\ve^m, \ve^c) + 1)}
\end{equation}

where, $\ve^k$ denotes flattened gradients of label $k$,  $q_i^k$ denotes soft label for the k-th label $k$ and $\alpha$ is a constant ($\alpha > 0.5$). 
The derived ($\vx_i, \vq_i$) pair is then used in the surrogate model training in addition to the true label $c$ (which is the only difference from the Train-ME).



\section{Model Extraction Performance} 


In this section, we demonstrate the performance of the proposed ME attacks and the baseline attack on SFL schemes.  All experiments are conducted on a single RTX-3090 GPU.
We use VGG-11 which has 11 layers as the model architecture. We vary $N$~(the the number of layers in server-side model) from 2 to 8 to generate different SFL schemes and evaluate them on CIFAR-10.

For the SFL model training, we set the total number of epochs to 200, and use SGD optimizer with a learning rate of 0.05 and learning rate decay (multiply by factor of 0.2 at epochs 60, 120 and 160). We assume all clients participate in every epoch with an equal number of training steps. We set the number of clients to 10 which corresponds to the cross-silo case. To perform ME attacks, the attacker uses an SGD optimizer with a learning rate of 0.02 to train the surrogate model and we report the best accuracy and fidelity. We evaluate accuracy of the surrogate model on the validation dataset.
We use the label agreement as fidelity, defined as the percentage of samples that the surrogate and victim models agree with over the entire validation dataset, as in \cite{jagielski2020high}. We include details of the SFL setting detail in~\cref{apx:setting}.

\subsection{ME Attack on SFL with Fine-tuning-based Training} 
\label{sec:fine-tune}

We first perform the proposed ME attacks on fine-tuning SFL version with consistent gradient query. Here we use a pre-trained model and set the number of gradient queries to 100K. On a victim VGG-11 model on CIFAR-10 dataset, whose original accuracy is 91.89\%, performance of all five ME attacks are shown in Table~\ref{tab:stable_brief}. For each of the ME attacks, we vary hyper-parameters and report the one that achieves the best attack performance. When $N=2$, all five ME attacks are successful and can achieve near-optimal accuracy and fidelity performance.
However, when $N$ is large, proposed ME attacks have worse attack performance.
For Craft, GAN and GM ME, the accuracy drops to around 80\% when $N$ is 5, and sharply drops to below 40\% when $N$ is 8. 
And for Train and SoftTrain ME, accuracy slightly degrades when $N$ is 5, and reduces to around 70\% when $N$ is 8. 
These results show that ME attack performance strongly correlates with $N$.  ME attack performance reduces for larger $N$ as the extraction problem becomes harder with more unknown parameters and more complicated input feature space.

\begin{figure}[htbp]
  \centering
  \includegraphics[width=1.0\linewidth]{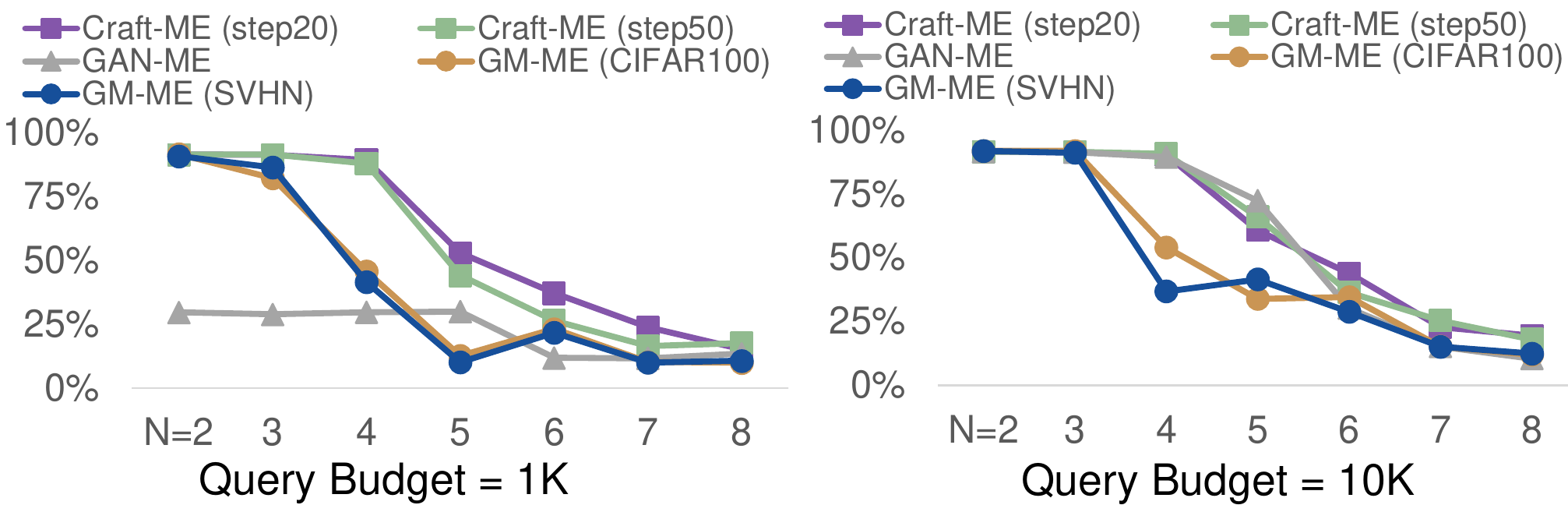}
\caption{ME attacks without training data under limited query budget (1K and 10K).}
\label{fig:query_budget}
\end{figure}

\textbf{Limited Query Budget.} Gradient queries are very important for our proposed attacks. We verify this by limit the qeury budget (original is 100K in Table~\ref{tab:stable_brief}). For Craft, GAN and GM MEs, we lower down the number of gradient queries to 1K and 10K. Results are shown in \cref{fig:query_budget}. Performance of all three ME attacks reduces significantly for a large N. But most of them (except GAN-ME) still achieves success when N is less than 4.

\textbf{Summary.} We summarize our findings based on results in Table~\ref{tab:stable_brief} and extensive evaluation in~\cref{apx:ideal_case}. 
In the fine-tuning case, we see that Craft, GAN and GM ME attacks are successful without training data. We observe the following interesting characteristics:
\textbf{Craft-ME} has a steady attack performance and can succeed even with a tight gradient query budget. 
\textbf{GAN-ME} needs a large query budget to train the c-GAN generator towards convergence but can achieve better accuracy and fidelity than Craft-ME for $N\leq5$. 
\textbf{GM-ME} requires an auxiliary dataset that is similar to the training data and when CIFAR-100 is used to attack CIFAR-10 model, GM-ME achieves almost perfect extraction for small $N$. However, it has slightly worse performance if MNIST or SVHN are used as auxiliary datasets. For large $N$, the surrogate model fails to converge on the GM loss, and its extraction performance suffers from a sharp drop. 
For attacks with training data such as \textbf{Train-ME} and \textbf{ SoftTrain-ME}, both accuracy and fidelity are much higher than attacks without training data. When $N\geq6$, SoftTrain-ME can achieve slightly better accuracy and fidelity than Train-ME.

\subsection{ME Attack on Training-from-scratch SFL}
\label{sec:practical}
Next, we investigate the proposed ME attack performance in training-from-scratch SFL case.
A good attack-time-window for gradient-based ME attacks is at the end of training when gradients do not vary as much and the model converges.
So for Craft, GAN and GM-ME, we launch the attack at epoch 160 to get more consistent gradients. As the model is updated by multiple clients, the percentage of malicious clients also affects the ME attack performance. We found that with more malicious clients, the server-side model returns more consistent gradients to the attacker.
Attack performance for three attacks are shown in Table~\ref{tab:stable_brief} for 10-client SFL training-from-scratch case.

\textbf{Summary.} We summarize our findings based on results in Table~\ref{tab:stable_brief} and extensive evaluations in~\cref{apx:practical_case}. 
In the training-from-scratch case, 
for gradient-based attacks without training data (Craft, GAN and GM MEs), we notice significant attack performance drop compared to the consistent query case. However they still can succeed in attacking a small server-side model ($N=5$) with around 86\% accuracy.
The same trend is shown in SoftTrain-ME. However SoftTrain's performance drops and that makes Train-ME the most effective attack for an attacker with training data.
We find that \textit{poison effect} and \textit{inconsistent gradients} contribute to the sharp drop in ME attack performance. Since gradient-based ME attacks require the attacker (as a participant) send noisy inputs (Craft, GAN and GM MEs) or genuine inputs with incorrect labels (SoftTrain-ME), the model accuracy suffers from a 2-3\% degradation, resulting in a less accurate target model. The inconsistent gradients reduces the effect of gradient query since model parameter can change rapidly. For instance, in Craft-ME, crafted inputs that have a small loss at an earlier epoch of the training can have a large loss in the final model because of the update of model parameters. Hence, inconsistent gradient information results in poor accuracy in the surrogate model.
Interestingly, compared to Craft-ME, GAN-ME is more robust to inconsistent gradients as the generator can adjust itself to the change of server-side model, resulting in a better attack performance when $N=5$. However, when $N$ is larger, the generator does not converge well and its performance drops drastically.
Last but not the least, GM-ME completely fails with inconsistent gradients, even for small $N$. This implies that the GM loss is super sensitive to inconsistent gradients and is only effective in consistent query cases.

\begin{figure}[htbp]
\centering
\begin{subfigure}{.49\columnwidth}
  \centering
  \includegraphics[width=1.0\linewidth]{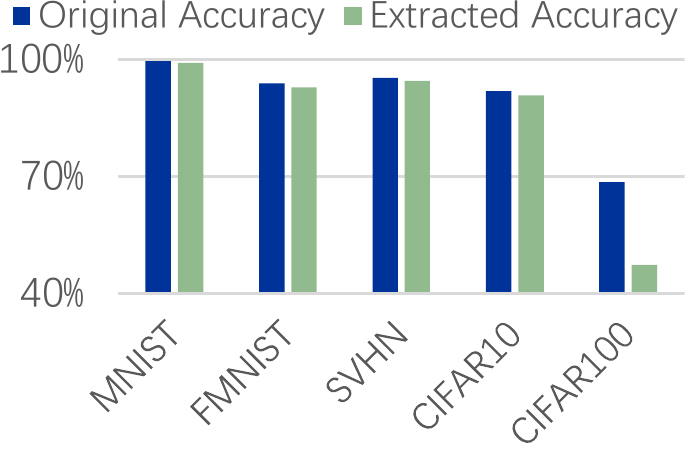}
  \label{fig:four_sub1}
\caption{}
\end{subfigure}%
\begin{subfigure}{.49\columnwidth}
  \centering
  \includegraphics[width=1.0\linewidth]{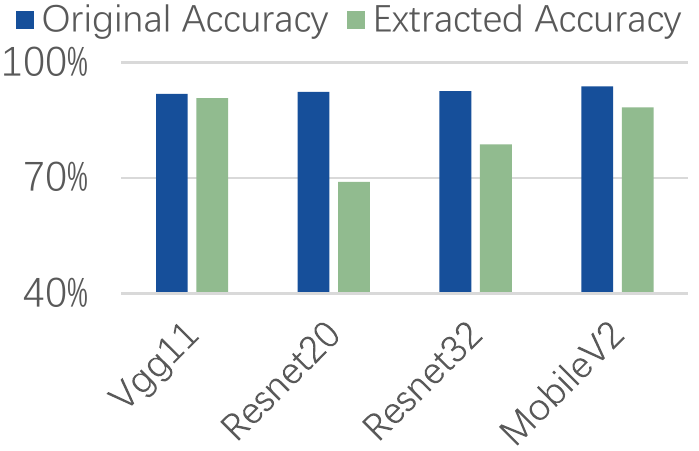}
  \label{fig:four_sub2}
\caption{}
\end{subfigure}
\caption{(a) ME attack performance of VGG-11 on other datasets. (b) ME attack performance of other architectures on CIFAR-10 dataset.}
\label{fig:four_figures}
\end{figure}

\subsection{ME Attack on SFL on other datasets and architectures} 
\label{sec:others}


First, we perform Train-ME attack with 1K training data on different datasets including MNIST~\cite{lecun1998mnist}, FMNSIT~\cite{xiao2017fashion}, SVHN~\cite{netzer2011reading} and CIFAR-100 datasets~\cite{krizhevsky2009learning}, using a VGG-11 model with $N$ set to 5. As shown in~\cref{fig:four_figures}~(a), for all datasets except CIFAR-100, ME attack achieves accuracy very close to the original. But for CIFAR-100, the extracted accuracy is low that is $>$20\% below the original.
Additionally, we also test Train-ME performance with 2\% and 20\% ImageNet training data on Mobilenet-V2.
As shown in~\cref{fig:tradeoff}~(a), ME attacks are hard to succeed due to the complexity of ImageNet dataset~\cite{deng2009imagenet}, resulting in a high accuracy gap of 10\% when $N$ is set to 2.
Second, we test Train-ME attack on different architectures including Resnet-20, Resnet-32~\cite{he2016deep} and Mobilenet-V2~\cite{sandler2018mobilenetv2} on CIFAR-10 dataset (with necessary adaptations)
For Resnet and Mobilenet family, we assign last 4 layer-blocks and 1 FC layer to server-side model. As shown in~\cref{fig:four_figures}~(b), with the same proportion of layers (5 out of 11) being assigned to server-side model, ME attack is much less effective on Resnet-20 than on VGG-11. A comparison of the performance of Resnet-32 and Mobilenetv2 with similar proportion of layers being assigned to server-side (5 out of 17 and 20, respectively), ME on Resnet-32 is also much worse than on MobilenetV2.

\textbf{Summary.} We find complex datasets such as CIFAR-100, ImageNet tend to be more resistant to ME attacks. Also, some architectures such as Resnet-20 and Mobilenet-V2 are more resistant than VGG-11.

\section{Discussion}
\label{sec:discussion}

\subsection{Tradeoff between IP and Data protection}
\begin{figure}[htbp]
  \centering
  
\begin{subfigure}{.46\columnwidth}
  \centering
  \includegraphics[width=1.0\linewidth]{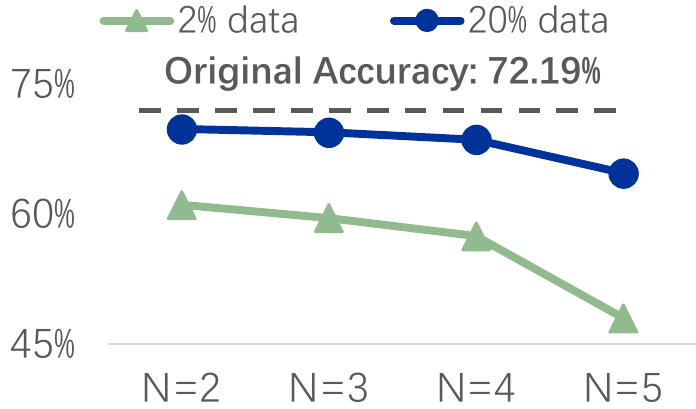}
  \label{fig:four_sub3}
\caption{}
\end{subfigure}
  \begin{subfigure}{.53\columnwidth}
  \centering
  \includegraphics[width=1.0\linewidth]{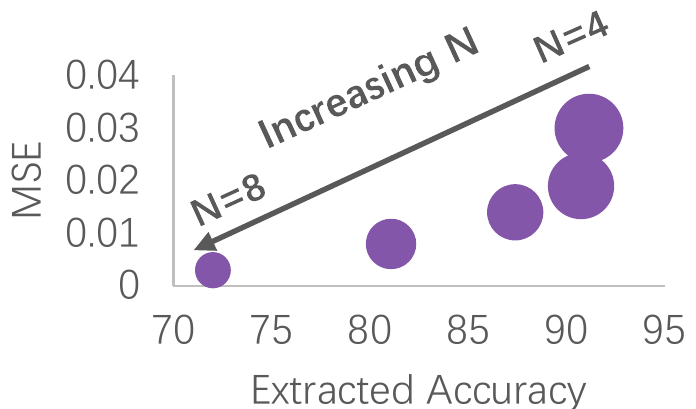}
    \label{fig:four_sub4}
    \caption{}
    \end{subfigure}
\caption{(a) ME attack performance of MobilenetV2 on ImageNet. (b) Tradeoff between ME resistance and degree of data protection (MSE).}
\label{fig:tradeoff}
\end{figure}

Our evaluation showed that ME attack performance drops with increasing $N$~(the number of layers in server-side model). Thus, a simple idea to improve resistance to ME attack is to use a larger $N$. However this implies that the number of layers in client-side model would be smaller,
thereby undermining clients' data. 
The tradeoff between IP protection and data protection is shown in~\cref{fig:tradeoff}~(b). We use Mean Square Error (MSE) of  reconstructed images by MI attack as a metric to represent the degree of data protection in SFL, as in \cite{li2022ressfl}; its implementation detail is included in ~\cref{apx:MI_attack}. 
From ~\cref{fig:tradeoff}~(b), we can see that with larger $N$, the extracted accuracy decreases  but MSE decreases meaning MI attack is more successful and clients' data protection is compromised. 
Thus, IP threat mitigation cannot be simply done by increasing $N$.

\subsection{Potential Defenses}

\begin{figure}[htbp]
  \centering
  \includegraphics[width=1.0\linewidth]{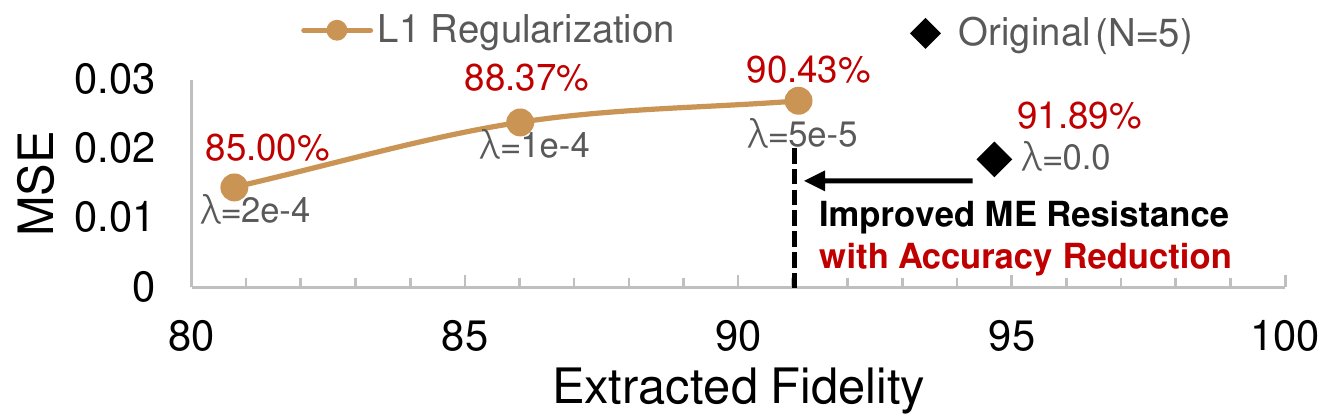}
\caption{L1 regularization as effective defense for ME attacks.}
\label{fig:regularization}
\end{figure}

Next we demonstrate how simple regularization can be used as defensive methods against ME attacks on SFL.
The key idea is to 
restrict the useful information in the client-side model that is leaked to the attack.
This is done by applying regularization techniques to restrict the client-side model's feature extraction capabilities. Specifically, on the client-side model, we apply L1 regularization with three different strength ($\lambda$ = 5e-5, 1e-4 and 2e-4) to penalize its weight magnitude.
As shown in~\cref{fig:regularization} for Train-ME with 1K data,
this simple defense effectively improves the resistance to ME attack though there is some accuracy degradation on the original model. For example, accuracy degrades to 90.43\% when $\lambda$ = 5e-5. Thus regularization can be used to defend ME attacks. More details are provided in~\cref{apx:impact_defenses}.

\subsection{Ablation Study}

\textbf{ME attack with non-IID data.} We consider the non-IID (independent and identically distributed) case where the attacker only has data from $C$ classes of CIFAR-10. Results presented in~\cref{apx:noniid} show that ME attack performance is still good for $C=5$ but degrades sharply when $C=2$.

\textbf{Adversarial attack based on successful ME attack.} The goal of ME attack is to launch more successful adversarial attacks. We perform transfer adversarial attacks using a surrogate model extracted by the strongest Train-ME attack. As shown in~\cref{apx:adversarial}, SFL with proper $N$ achieves better resistance to adversarial attacks.

\textbf{ME attack without architecture information.} In~\cref{apx:diff_arch}, we investigate simple variants (longer, shorter, wider, and thinner) of the original architecture as the surrogate model architecture. We find that the performance of ME attacks is similar for the different architectures -- the exception is GM-ME which fails for different surrogate architectures.

\section{Conclusion}
In this work, we show that SFL cannot guarantee model IP protection and is vulnerable to ME attacks.
We propose five novel ME attack methods and achieve attack success under an unique threat model where gradient query is allowed but prediction query is not allowed.  
By studying the effect of the model split sizes on ME attack performance, we find that using a large  number of layers in server-side model can better protect IP, but compromises clients' data and is hence impractical. 
Finally, as a first step towards making SFL resistant to ME attack, we show use of regularization as a potential defense mechanism.


{\small
\bibliographystyle{ieee_fullname}
\bibliography{IEEEtran}
}
\clearpage
\appendix
\noindent\textbf{Supplementary Material}
\section{Details of Experimental Settings and Additional Results}

\subsection{Detailed Extraction Attack Setting}
\label{apx:setting}
For the surrogate model training, we use SGD optimizer with a learning rate of 0.02 for 200 epochs. The learning rate is multiplied by a factor of 0.2 at epochs 60, 120 and 160.
For Craft-ME, we craft an equal number of small-loss instances  for each class. We use the Adam optimizer with a learning rate of 0.1 and set the total number of steps~(iterations) to 20 or 50 to craft each image. For GAN-ME, we use a conditional-GAN model as the generator, its detailed architecture is given in Appendix~\ref{sec:cgan}. For the generator training, we use Adam optimizer with a learning rate of 1e-4 and apply the divergence-aware regularization \cite{yang2019diversity} with a factor of 50 to mitigate the mode collapse problem. For GM-ME, we query with the entire training dataset (i.e. 50K) of each auxiliary dataset (CIFAR-10, SVHN and MNIST). For Train-ME and SoftTrain-ME, we apply standard data augmentation techniques including random rotating and horizontal flipping, during the surrogate model training. For SoftTrain-ME, we train the surrogate model with both the hard labels and gradient-based soft labels with $\alpha$ parameter of 0.9. Data augmentation is disabled if training uses soft labels. 





\subsection{Gradient-based Attack Performance with Consistent Gradient Access}
\label{apx:ideal_case}
\begin{figure*}[htbp]
\centering
\begin{subfigure}{.33\textwidth}
  \centering
  \includegraphics[width=0.99\linewidth]{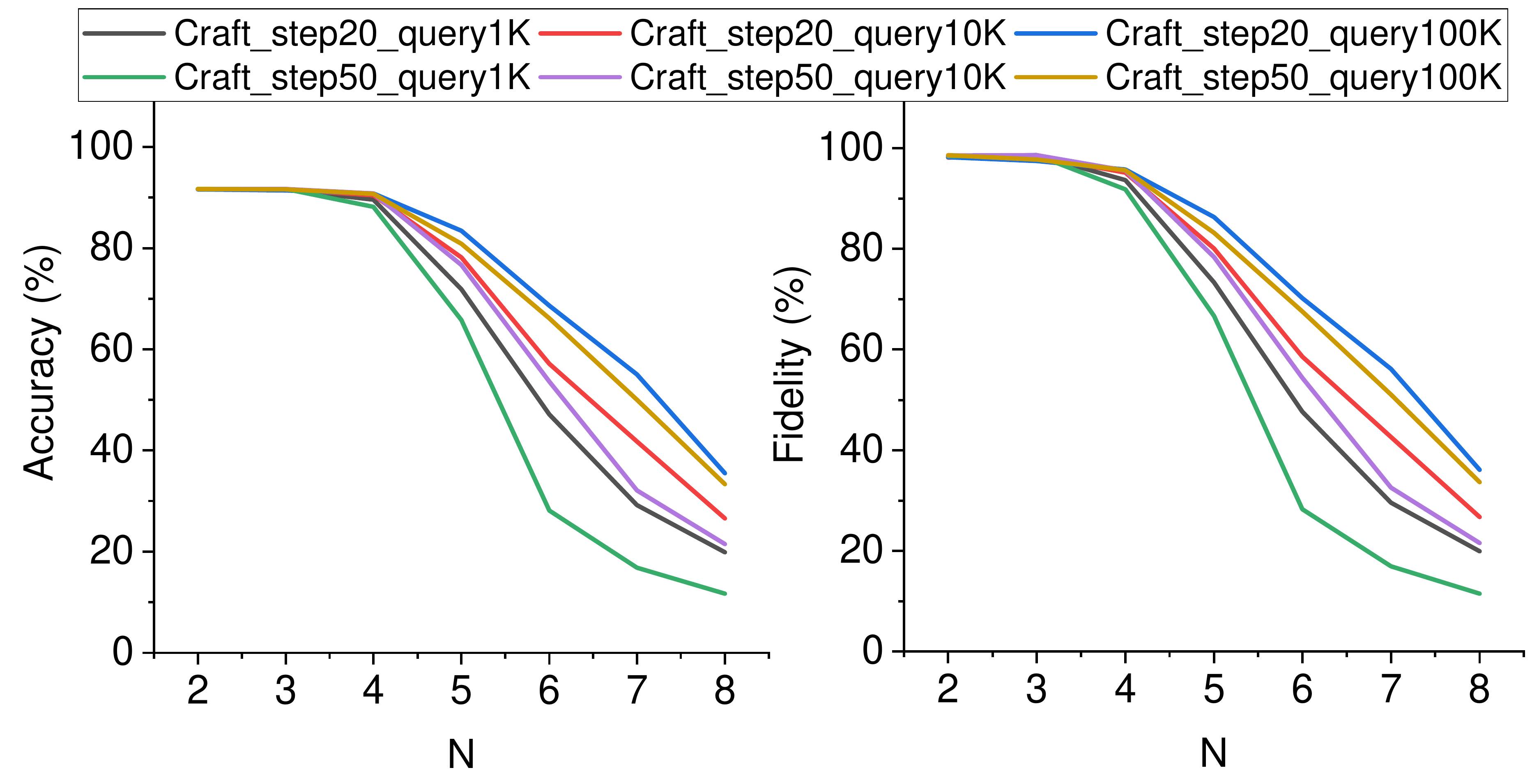}
  \label{fig:stable_sub1}
\vspace{-1.0em}
\caption{}
\end{subfigure}%
\begin{subfigure}{.33\textwidth}
  \centering
  \includegraphics[width=0.99\linewidth]{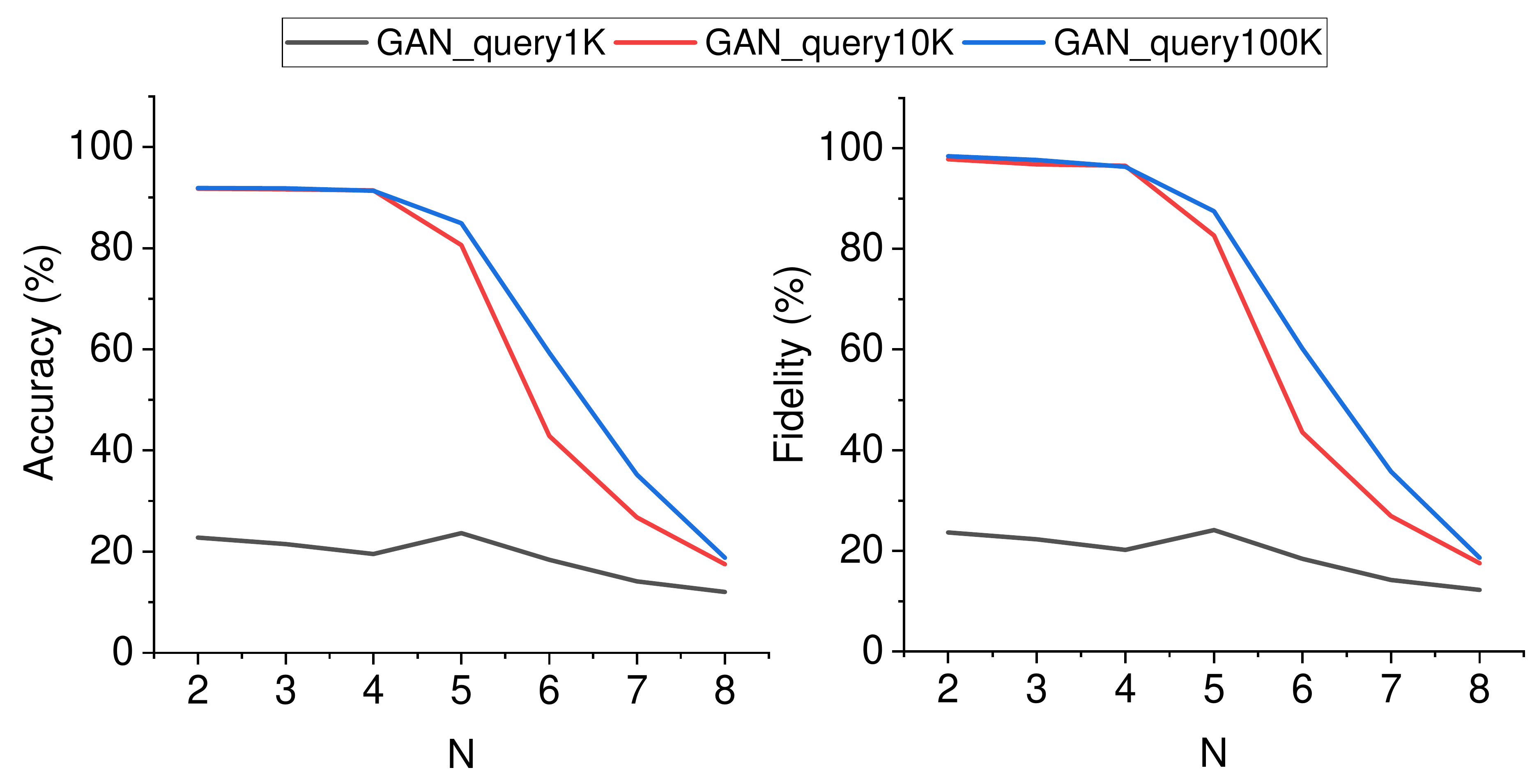}
  \label{fig:stable_sub2}
\vspace{-1.0em}
\caption{}
\end{subfigure}
\begin{subfigure}{.33\textwidth}
  \centering
  \includegraphics[width=0.99\linewidth]{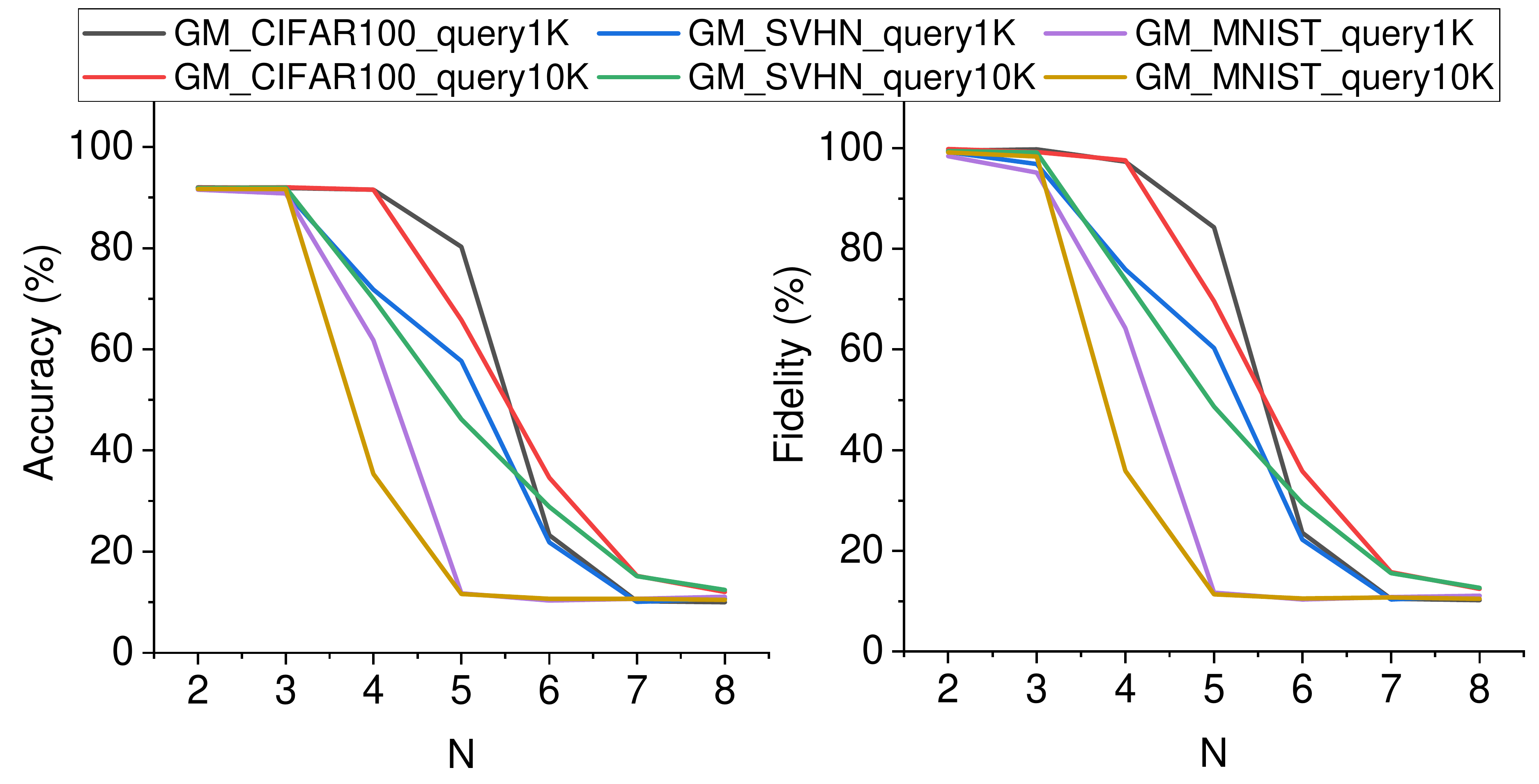}
  \label{fig:stable_sub3}
\vspace{-1.0em}
\caption{}
\end{subfigure}
\begin{subfigure}{.49\textwidth}
  \centering
  \includegraphics[width=0.99\linewidth]{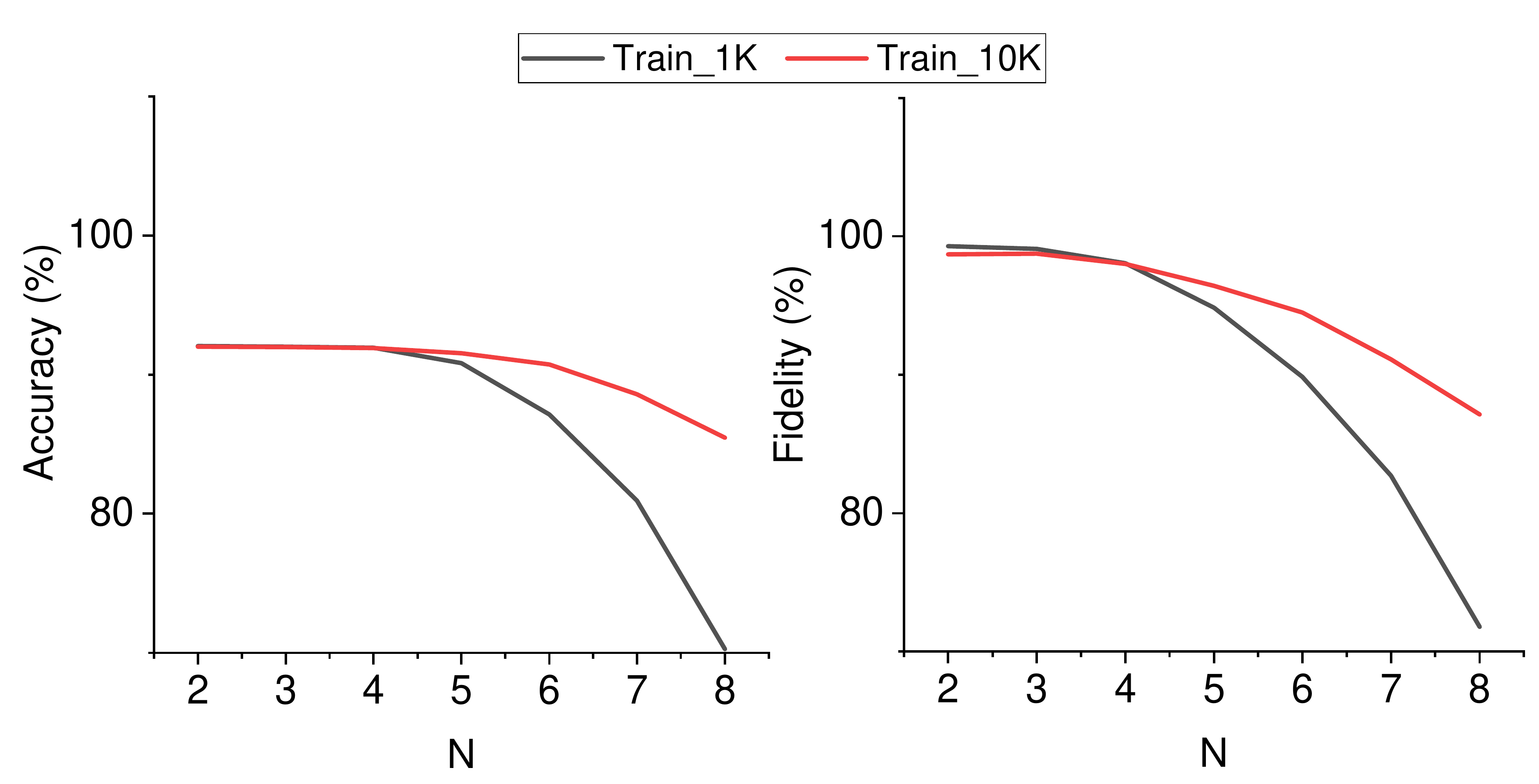}
  \label{fig:stable_sub4}
\vspace{-0.5em}
\caption{}
\end{subfigure}
\begin{subfigure}{.49\textwidth}
  \centering
  \includegraphics[width=0.99\linewidth]{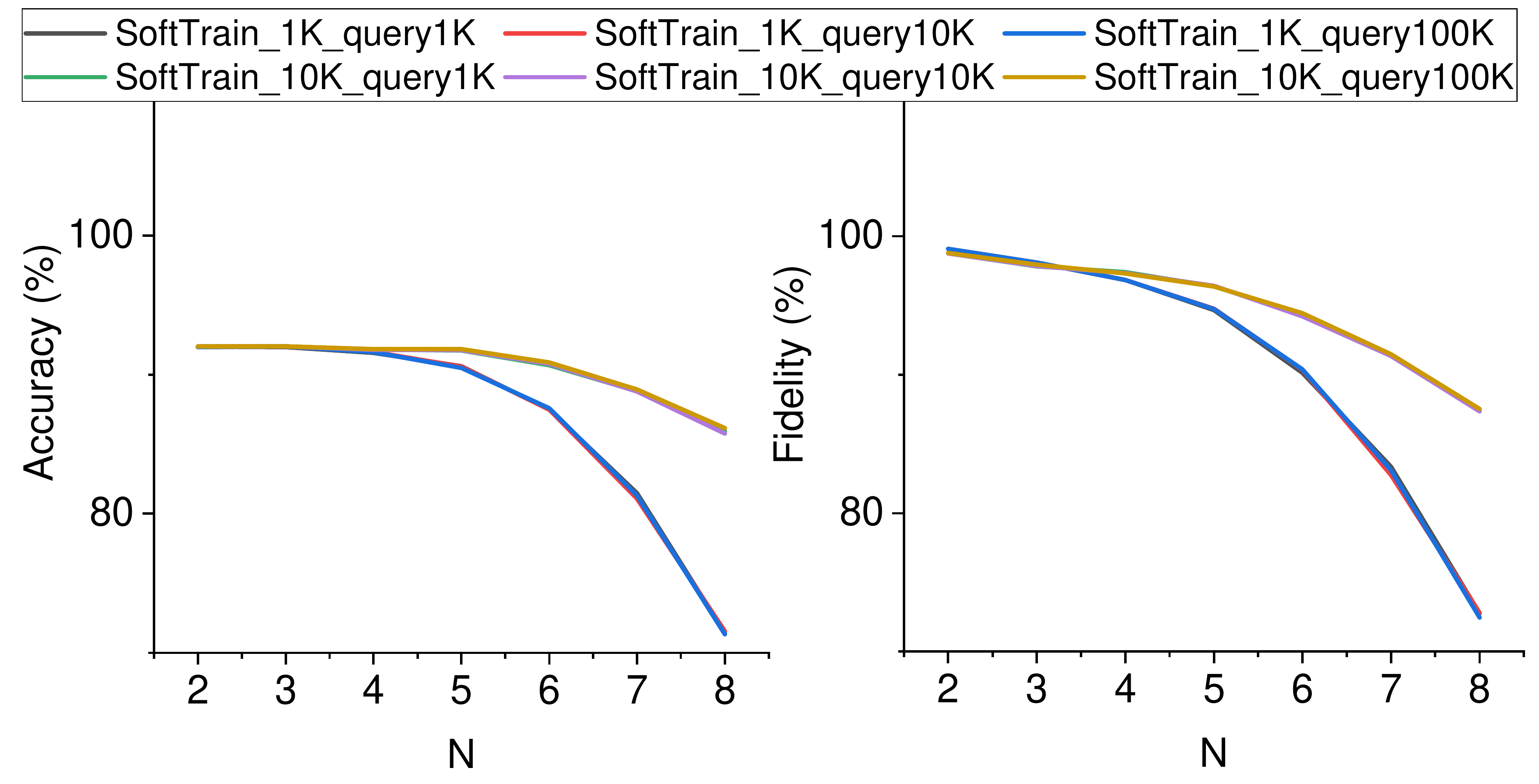}
  \label{fig:stable_sub5}
\vspace{-0.5em}
\caption{}
\end{subfigure}
\caption{Additional results for \textbf{fine-tuning} SFL case. \textbf{Top row:} ME attacks without training data with different settings. (a) Craft-ME with different number of crafting steps and query budgets. (b) GAN-ME with different query budgets. (c) GM-ME with different auxiliary datasets and query budgets. 
\textbf{Bottom row:} ME attacks with training data with different settings. (d) Train-ME with 1K/10K training data. (e) SoftTrain-ME with 1K/10K training data with different query budgets.}
\vspace{-0.5em}
\label{fig:stable_figures}
\vspace{-0.5em}
\end{figure*}

We did extensive experiments for ME attacks in different settings with consistent gradient query access and present the results here. These are in addition to what was presented in Section~\ref{sec:fine-tune}. The query budget is set at 1K, 10K and 100K. Results for all five ME attacks with different settings are shown separately in~\cref{fig:stable_figures} (a), (b), (c), (d) and (e). The victim VGG-11 model has 91.89\% validation accuracy on CIFAR-10 dataset.
For GM-ME, we CIFAR-100, SVHN, and MNIST as the auxiliary dataset. 

\textbf{Conclusion.} We observe all ME attacks are equally successful for small $N$. Among different settings, Craft-ME performs better with 20 steps compared to 50 steps. This is possibly because for the same query budget, fewer steps results in more images being crafted.
GAN-ME performance is much better with a larger query budget since a generator model needs more iterations of training to converge. GM-ME's performance heavily depends on the similarity of the auxiliary dataset. Because the victim model is on CIFAR-10, it performs well when CIFAR-100 is set as the auxiliary dataset while performing badly when MNIST is used. Moreover, attacks with training data perform much better than ME attacks without training data. Compared to Train-ME, SoftTrain-ME achieves better accuracy and fidelity when $N\geq6$.

\subsection{Gradient-based Attack Performance with Inconsistent Gradient Access}
\label{apx:practical_case}
Here, we provide results for ME attacks for training-from-scratch settings with inconsistent gradient query access; a subset of these results was presented in Section~\ref{sec:practical}.
We launch the attack by feeding malicious inputs at late epochs, specifically, epochs 120 or 160, for the case when the number of training epochs is 200. The attacker starts to collect gradients after the attack is launched till the the end of training (epoch 200). 
We start the gradient collection from later epochs since by then the model has achieved near-optimal accuracy and hence is valuable as an attack target.  Also the model updating is slower because of application of learning rate decay to make the gradients more consistent.

\begin{figure*}[htbp]
\centering
\begin{subfigure}{.33\textwidth}
  \centering
  \includegraphics[width=0.99\linewidth]{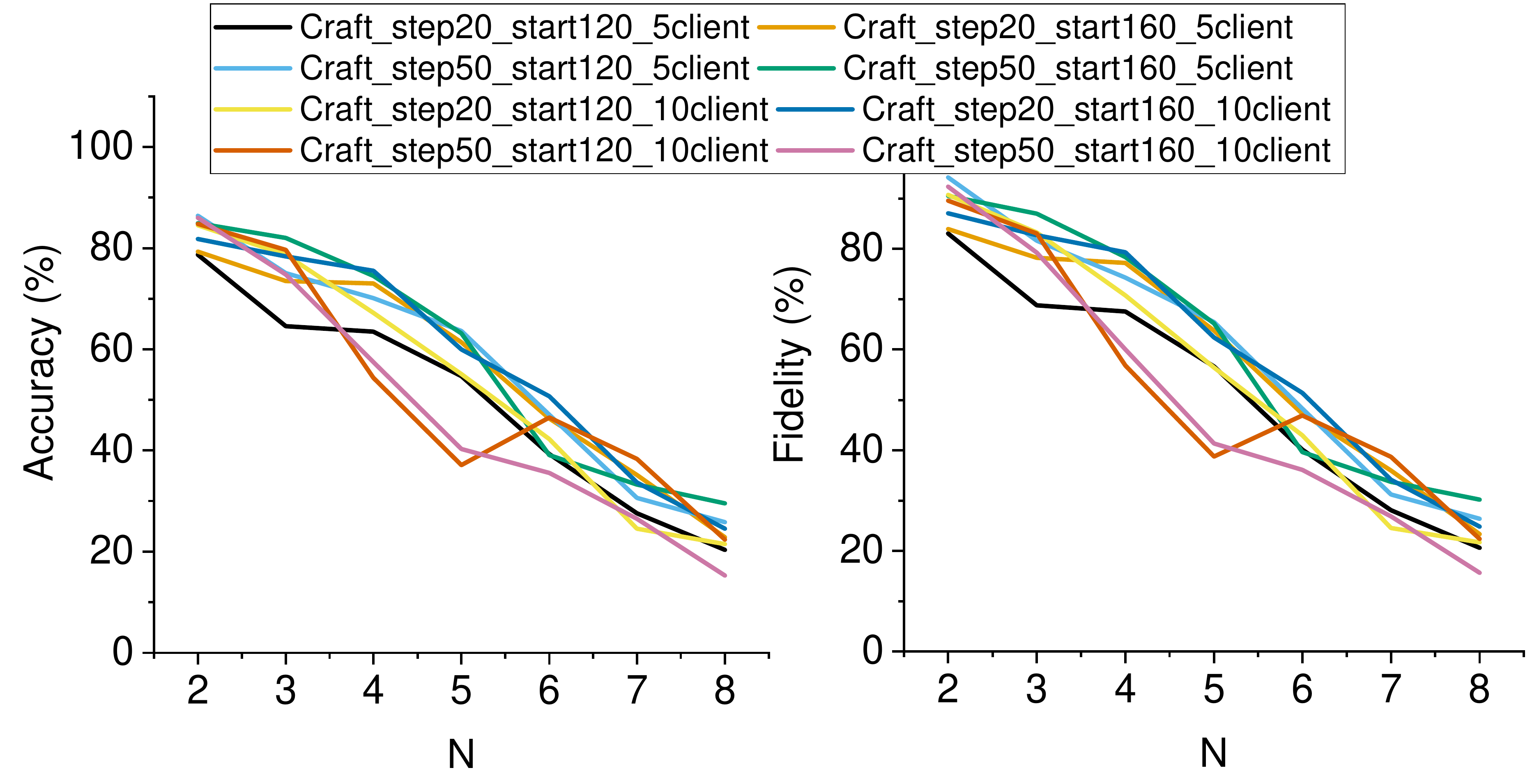}
  \label{fig:unstable_sub1}
\vspace{-1.0em}
\caption{}
\end{subfigure}%
\begin{subfigure}{.33\textwidth}
  \centering
  \includegraphics[width=0.99\linewidth]{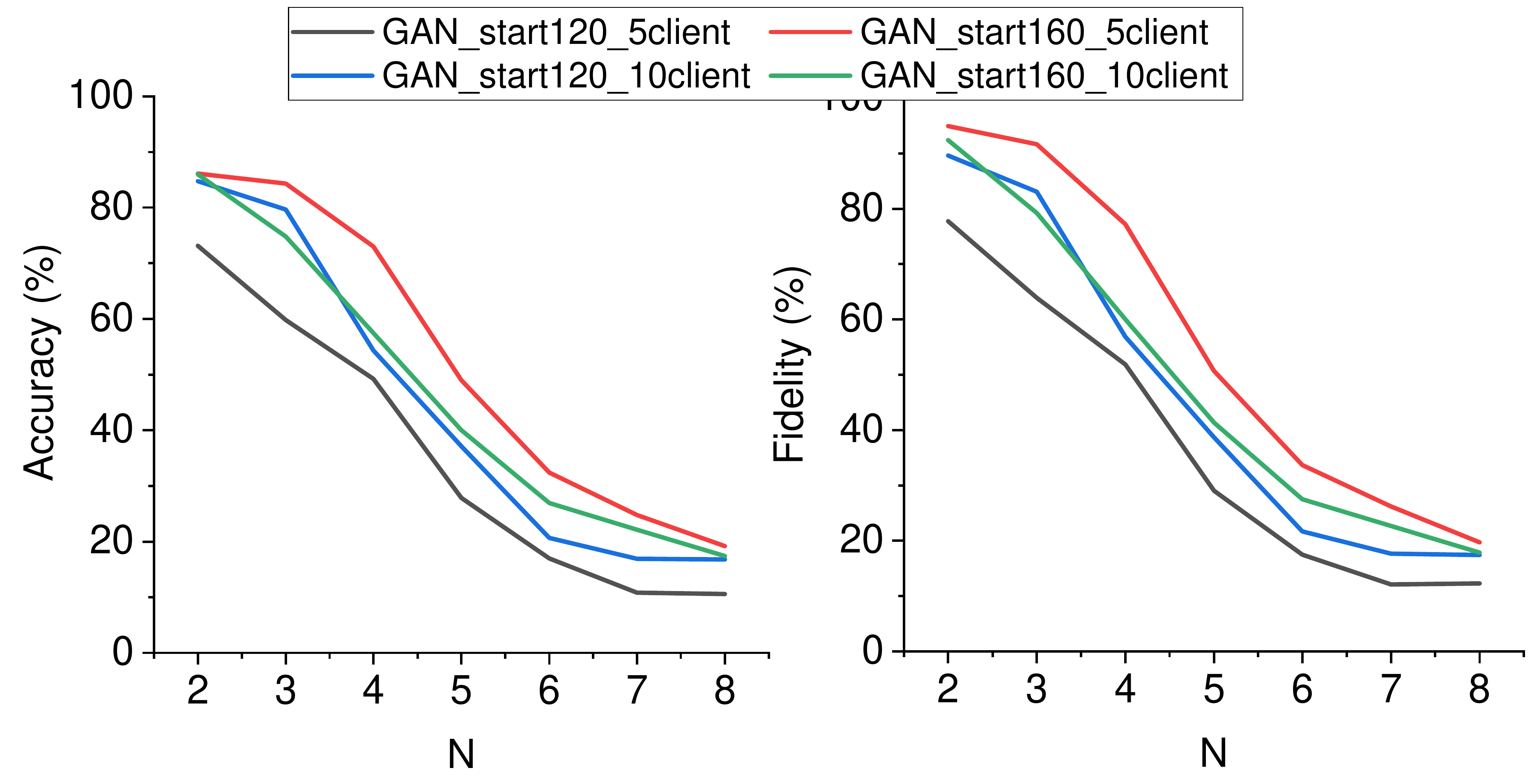}
  \label{fig:unstable_sub2}
\vspace{-1.0em}
\caption{}
\end{subfigure}
\begin{subfigure}{.33\textwidth}
  \centering
  \includegraphics[width=0.99\linewidth]{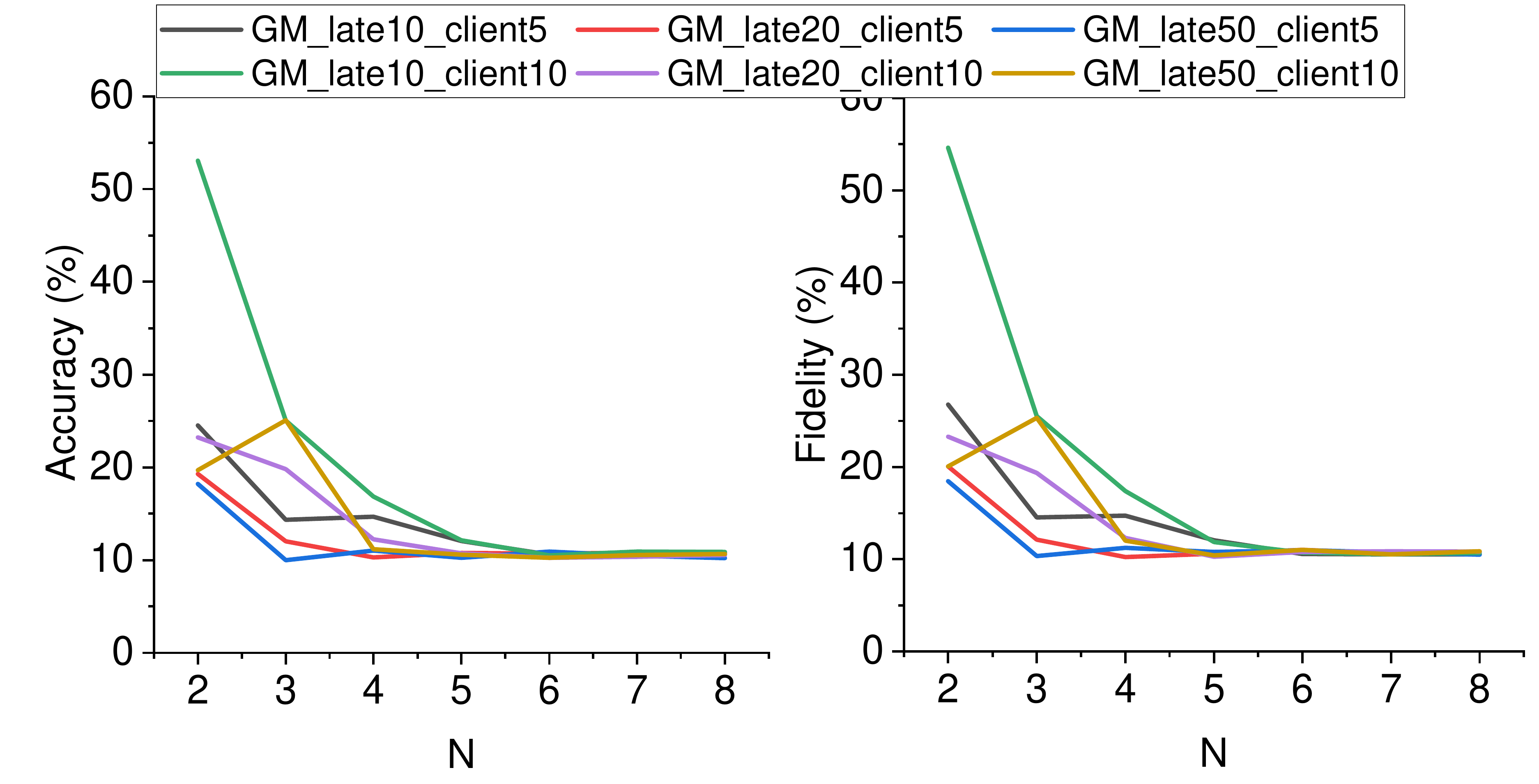}
  \label{fig:unstable_sub3}
\vspace{-1.0em}
\caption{}
\end{subfigure}
\begin{subfigure}{.49\textwidth}
  \centering
  \includegraphics[width=0.99\linewidth]{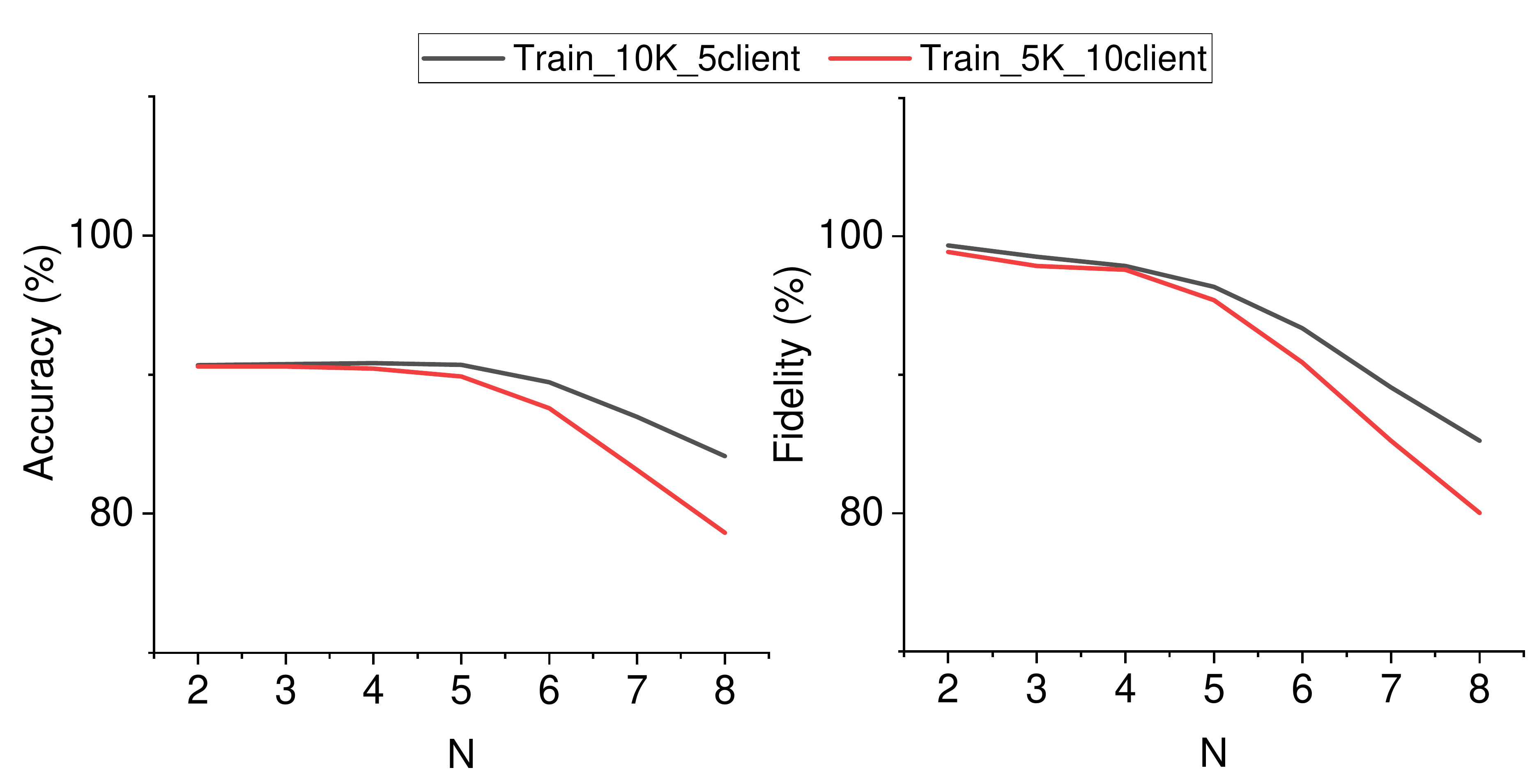}
  \label{fig:unstable_sub4}
\vspace{-0.5em}
\caption{}
\end{subfigure}
\begin{subfigure}{.49\textwidth}
  \centering
  \includegraphics[width=0.99\linewidth]{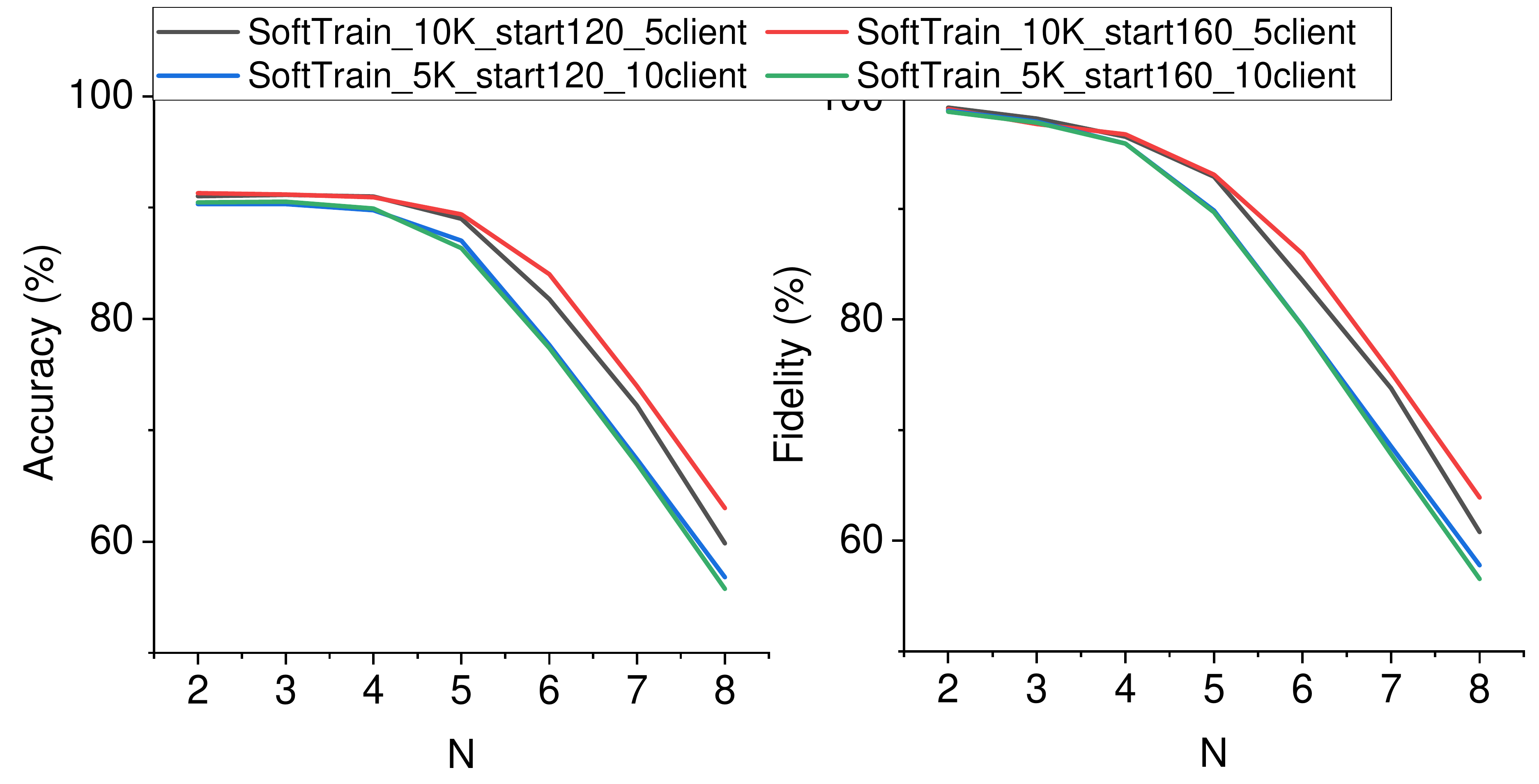}
  \label{fig:unstable_sub5}
\vspace{-0.5em}
\caption{}
\end{subfigure}
\caption{Additional results for \textbf{training-from-scratch} SFL case. \textbf{Top row:} ME attacks without training data with different settings. (a) Craft-ME with different steps and starting epochs. (b) GAN-ME with different starting epochs. (c) GM-ME uses the latest gradients with different restrictions. 
\textbf{Bottom row:} ME attacks with training data with different settings. (d) Train-ME with 10K/5K training data. (e) SoftTrain-ME with 10K/5K training data with different starting epochs.}
\vspace{-0.5em}
\label{fig:unstable_figures}
\vspace{-0.5em}
\end{figure*}

In multi-client SFL, the original 50K training data is divided equally to 5 or 10 benign clients, denoted as ``5-client'' and ``10-client'' case, respectively. The attacker is an additional client without training data so a 5-client SFL really has 6 clients (5 benign clients and 1 malicious client). All clients, including the attacker, perform an equal number of queries in each epoch.
The performance of five ME attacks with inconsistent gradient queries are shown separately in~\cref{fig:unstable_figures}~(a), (b), (c), (d) and (e). Because of the poisoning effect, the final model accuracy of the victim model is reduced by 2 $\sim$ 3\%. For GM-ME, we use CIFAR-100 as the auxiliary dataset, and we only use the latest gradients to perform  gradient matching instead of using all collected gradients. We use ``late50'' to denote only gradients collected in 50 latest training steps are used. This restriction greatly reduces the number of gradients being available but makes them much more consistent.

\textbf{Conclusion.} Attacks without training data (Craft, GAN, GM MEs) work poorly with inconsistent gradient queries. For Craft-ME, taking 20 steps also seems to work better in both 5-client and 10-client cases. Collecting gradients starting later at epoch 160 gets better performance than starting early at epoch 120  because of more consistent gradients. The starting-later rule also holds for GAN-ME and GM-ME, where we can see starting later achieves consistently better ME attack performance. For GM-ME, it only gets meaningful accuracy if only the latest gradients (within 10 training steps to the end of training) are used, showing that it is extremely sensitive to gradient consistency. For attacks with training data, we notice Train-ME attack performance is not affected because it does not rely on gradients. However, SoftTrain-ME performs much worse because of the poisoning effect and inconsistent gradients.

\subsection{Accuracy Impact of Defensive Methods}
\label{apx:impact_defenses}

We provide original accuracy, ME attack performance, as well as model inversion attack performance in addition to what was presented in the main paper in section~\ref{sec:discussion}. As shown in Table~\ref{tab:defense_detail},
L1 regularization works well for $N=5$ where it reduces extraction performance a lot while slightly affecting original accuracy, and at the same time also improves the resistance to model inversion attack (better data protection). 

\begin{table*}[htbp]
\caption{Detailed defensive performance of L1 regularization (L1Reg) of VGG-11 model on CIFAR-10. Extraction performance of Train-ME with 1K training data is shown. Resistance to model inversion attack is shown by MSE.}
\label{tab:defense_detail}
\begin{center}
\begin{small}
\resizebox{1.0\linewidth}{!}{
\begin{tabular}{lc|cccc|cccc} 
 \toprule
 \multirow{2}{*}{Regularization}& \multirow{2}{*}{Strength}&\multicolumn{4}{c}{\textbf{N=4}} & \multicolumn{4}{c}{\textbf{N=5}}\\
  & &Orig. Accu. & Accuracy & Fidelity & MSE &Orig. Accu. & Accuracy & Fidelity & MSE\\
 \midrule
 None &0.0 &91.45&91.02&96.94&0.0217&91.71&90.23&94.73&0.0114\\
 L1Reg &5e-5 &90.66&89.44&95.03&0.0274&90.43&87.45&91.10&0.0270\\
 L1Reg &1e-4 &87.90&86.24&93.68&0.0280&88.37&82.18&86.01&0.0239\\
 L1Reg &2e-4 &82.96&80.56&89.98&0.0262&85.00&76.45&80.78&0.0145\\
 \bottomrule
\end{tabular}
}
\end{small}
\end{center}
\end{table*}



\subsection{Extraction Performance under Non-IID data distribution}
\label{apx:noniid}

We demonstrate the  ME attack performance in a Non-IID setting, where the attacker only has access to training data from a subset of classes ($C$).
The new set of results corresponding to Train-ME attack are shown in Table~\ref{tab:noniid_brief}. We observe for $C$ smaller than 5, attacker performance degrades badly for both CIFAR-10 and CIFAR-100 datasets.

\begin{table}[htbp]
\caption{Under Non-IID data distribution, model extraction performance of Train-ME attack on VGG-11 model on CIFAR-10 and CIFAR-100 dataset with Original Accuracy of 91.89\% and 68.64\%, respectively.}
\label{tab:noniid_brief}
\begin{center}
\begin{small}
\resizebox{1.0\linewidth}{!}{
\begin{tabular}{lcccccc} 
 \toprule
 \multirow{2}{*}{Method} & \multicolumn{3}{c}{\textbf{CIFAR-10 Accuracy}} & \multicolumn{3}{c}{\textbf{CIFAR-100 Accuracy}}\\
  &N=2 &N=5 &N=8 &N=2 &N=5 &N=8\\
 \midrule
  C = 1  &47.58	&46.75	&38.42	&6.79	&6.79	&6.13\\
  C = 2  &82.45	&79.30	&58.69	&13.18	&13.36	&11.25\\
  C = 5  &91.70	&88.90	&65.70	&32.77	&29.65	&17.90\\
 \bottomrule
\end{tabular}
}
\end{small}
\end{center}
\end{table}

\subsection{Adversarial Attack Performance}
\label{apx:adversarial}
We demonstrate that with proper model IP protection, adversarial attacks can be mitigated. We assume the attacker uses the strongest Train-ME attack (with 1K data) to obtain a high-fidelity surrogate model to perform transfer adversarial attacks on the victim model with different IP protection strengths (SFL with different $N$). We use FGSM \cite{goodfellow2014explaining}, and targeted-PGD attack \cite{madry2017towards} to perform the transfer adversarial attack. We set the $e$ for FGSM at 0.1, and PGD-target at 0.002 for 50 iterations (the attacker randomly chooses the original and target label). We report the average Attack Success Rate (ASR) - the percentage of samples that are transferred successfully - to show the attacking performance. The new set of results is shown in Table~\ref{tab:advattack_brief}. We see that both adversarial attacks achieve very high ASR for small $N$, where model IP protection is weak. On a SFL scheme with large $N$, adversarial attack performance degrades significantly using the surrogate model with less fidelity.

\begin{table}[htbp]
\caption{Adversarial Attack ASR performance based on the surrogate model obtained using Train-ME attack, on VGG-11 model on CIFAR-10 with different $N$ setting.}
\label{tab:advattack_brief}
\begin{center}
\begin{small}
\resizebox{1.0\linewidth}{!}{
\begin{tabular}{lccccccc} 
 \toprule
 \multirow{2}{*}{Attack} & \multicolumn{7}{c}{Number of Server-side Layer ($N$)}\\
  &N=2 &N=3 &N=4 &N=5 &N=6 &N=7 &N=8\\
 \midrule
  FGSM  &82.7	&82.3	&77.9	&77.3	&63.1	&56.9 &37.7\\
  PGD-target& 100	&100	&99.8	&100	&99.5	&73.4 &34.2\\
 \bottomrule
\end{tabular}
}
\end{small}
\end{center}
\end{table}

\subsection{Surrogate Architecture Performance}
\label{apx:diff_arch}

To investigate the impact on model extraction attacks caused by the surrogate model's architecture difference, we designed four variants of the true server-side model, and used them as surrogate model architecture to perform model extraction attacks. We fixed the settings to  $N = 5$ SFL and  consistent gradient query budget to 10K. The new set of results are shown in Table~\ref{tab:diff_surrogate_arch} for VGG-11 model on CIFAR-10 dataset.
For most attacks, architecture does not make a huge difference, and longer or wider surrogate architecture can achieve even better accuracy and fidelity. The exception is GM-ME, which achieves much higher extraction performance with the surrogate model having the same architecture.

\textbf{Longer Architecture.}
Surrogate model has one extra fully connected layer compared to the original true server-side model.

\textbf{Shorter Architecture.}
Surrogate model has one less fully connected layer compared to the original true server-side model.

\textbf{Wider Architecture.}
Surrogate model has channel size that is 2 times of the original true server-side model

\textbf{Thinner Architecture.}
Surrogate model has channel size half of the original channel size of the true server-side model.

\begin{table*}[htbp]
\caption{Extraction attack performance on surrogate models having slightly different architectures from the true architecture of the server-side model. $N$ is fixed at 5, gradients are consistent and the query budget is 10K.}
\label{tab:diff_surrogate_arch}
\begin{center}
\begin{small}
\resizebox{0.9\linewidth}{!}{
\begin{tabular}{lccccc|ccccc} 
 \toprule
 \multirow{2}{*}{Attacks} & \multicolumn{5}{c}{\textbf{Accuracy (\%)}} & \multicolumn{5}{c}{\textbf{Fidelity (\%)}}\\
 &same &longer &shorter &wider &thinner &same &longer &shorter &wider &thinner\\
 \midrule
 Craft-ME &76.67 &75.05 &77.90	&79.00	&74.86	&78.38 &76.70	&79.72	&81.04 &74.74\\
 GAN-ME&80.57 &75.95   &76.66	&74.27	&65.69	&82.66 &78.13	&78.54	&76.11 &67.58\\
 GM-ME&65.77 &11.41    &18.04	&14.77	&14.42	&69.60 &11.22	&18.61	&14.90 &14.35\\
 Train-ME& 90.82 &90.33    &90.76	&90.72	&90.10	&94.84 &94.47	&94.84	&94.79 &93.94\\
 SoftTrain-ME&90.57&90.43   &90.66	&90.62	&90.10	&94.76 &94.62	&94.84	&94.59 &94.13\\
 \bottomrule
\end{tabular}
}
\end{small}
\end{center}
\end{table*}

\begin{table*}[htbp]
\caption{Model extraction performance of gradient-based ME attacks with consistent gradient query (100K query budget) and inconsistent gradient query for 10-client SFL on \textbf{VGG-11 model CIFAR-100 dataset}. Original Accuracy is 68.64\%. We use 20 crafting steps for the Craft-ME for both cases. For the inconsistent case, we launch ME attack at epoch 160, and use the ``late10'' setting for GM-ME.}
\label{tab:other_vgg11_cifar100}
\begin{center}
\begin{small}
\resizebox{0.8\linewidth}{!}{
\begin{tabular}{clcccc|cccc} 
 \toprule
 \multirow{2}{*}{Case} & \multirow{2}{*}{Method} & \multicolumn{4}{c}{\textbf{Accuracy (\%)}} & \multicolumn{4}{c}{\textbf{Fidelity (\%)}}\\
  & &N=2 &N=3 &N=4 &N=5 &N=2 &N=3 &N=4 &N=5\\
 \midrule
  \multirow{3}{*}{\makecell{Fine-tuning}} &Craft-ME &66.44	&64.68	&35.37	&15.4	&86.97	&81.37	&40.35	&16.7\\
  &GAN-ME   &56.54	&46.53	&13.11	&6.69	&69.91	&55.56	&14.86	&7.11\\
  &GM-ME    &68.76	&68.4	&57.87	&1.28	&99.11	&94.46	&71.5	&1.26\\
 \midrule
  \multirow{3}{*}{\makecell{Train-from-scratch}} &Craft-ME &11.53	&8.49	&2.61	&2.41	&13.67	&10.15	&2.71	&2.45\\
  &GAN-ME   &49.4	&41.9	&22.1	&10.75	&60.04	&49.29	&25.46	&12.55\\
  &GM-ME    &4.05	&1.47	&1.37	&1.23	&4.92	&1.79	&1.54	&1.19\\
 \bottomrule
\end{tabular}
}
\end{small}
\end{center}
\end{table*}

\begin{table*}[htbp]
\caption{Model extraction performance of gradient-based ME attacks with consistent gradient query (100K query budget) on \textbf{VGG-11 model FEMNIST dataset}. Original Accuracy is 74.62\%. We use 50 crafting steps for the Craft-ME for both cases.}
\label{tab:other_vgg11_femnist}
\begin{center}
\begin{small}
\resizebox{0.8\linewidth}{!}{
\begin{tabular}{clcccc|cccc} 
 \toprule
 \multirow{2}{*}{Case} & \multirow{2}{*}{Method} & \multicolumn{4}{c}{\textbf{Accuracy (\%)}} & \multicolumn{4}{c}{\textbf{Fidelity (\%)}}\\
  & &N=2 &N=3 &N=4 &N=5 &N=2 &N=3 &N=4 &N=5\\
 \midrule
  \multirow{5}{*}{\makecell{Fine-tuning}} &Craft-ME &53.20	&43.57	&43.04	&40.50	&59.28	&48.10	&46.58	&42.11\\
  &GAN-ME   &10.59	&7.19	&5.27	&4.02	&11.39	&7.35	&5.20	&3.80\\
  &GM-ME    &56.67	&22.53	&9.87	&3.78	&69.70	&25.14	&10.10	&3.68\\
  &Train-ME   &70.32 &68.47   &68.80  &67.70	&82.56	&77.52	&75.61	&71.97\\
  &SoftTrain-ME    &75.70	&74.93	&74.42	&74.46	&83.87	&81.24	&77.39	&76.30\\
 \bottomrule
\end{tabular}
}
\end{small}
\end{center}
\end{table*}

\begin{table*}[htbp]
\caption{Model extraction performance of gradient-based ME attacks with consistent gradient query (100K query budget) and inconsistent gradient query for 10-client SFL on \textbf{MobilenetV2 model CIFAR-10 dataset}. Original Accuracy is 93.82\%. We use 20 crafting steps for the Craft-ME for both cases. For the inconsistent case, we launch ME attack at epoch 160, and use the ``late10'' setting for GM-ME.}
\label{tab:other_mob_cifar10}
\begin{center}
\begin{small}
\resizebox{0.8\linewidth}{!}{
\begin{tabular}{clcccc|cccc} 
 \toprule
 \multirow{2}{*}{Case} & \multirow{2}{*}{Method} & \multicolumn{4}{c}{\textbf{Accuracy (\%)}} & \multicolumn{4}{c}{\textbf{Fidelity (\%)}}\\
  & &N=2 &N=3 &N=4 &N=5 &N=2 &N=3 &N=4 &N=5\\
 \midrule
  \multirow{3}{*}{\makecell{Fine-tuning}} &Craft-ME &92.29	&76.74	&72.74	&61.08	&96.04	&77.86	&73.24	&61.46\\
  &GAN-ME   &92.67	&79.17	&68.92	&57.61	&96.46	&80.35	&69.7	&58.25\\
  &GM-ME    &93.2	&92.82	&92.39	&91.86	&97.83	&96.87	&95.74	&94.74\\
 
 \midrule
  \multirow{3}{*}{\makecell{Train-from-scratch}} &Craft-ME &78.55	&63.04	&61.77	&58.76	&80.8	&64.87	&63.35	&60.07\\
  &GAN-ME   &77.08	&38.2	&35.08	&32.49	&79.43	&38.87	&35.6	&33.11\\
  &GM-ME    &31.23	&11.25	&15.25	&17.81	&32.73	&11.46	&15.14	&17.9\\
 \bottomrule
\end{tabular}
}
\end{small}
\end{center}
\end{table*}

\subsection{Other Empirical Results}
\label{apx:other_empirical}
In this section, we present more empirical results for ME attacks without training data to show that our claims can generalize to other architecture and datasets. The list of experiments are:

\begin{itemize}
    \item 1. ME attack performance (without training data only) of VGG-11 on CIFAR-100 (Table~\ref{tab:other_vgg11_cifar100}). An interesting observation is GAN-ME performs worse than Craft-ME for consistent gradient cases for the increasing number of classes (100) makes the generator even harder to converge. While for the inconsistent gradient case, GAN-ME performs much better than Craft-ME because its generator can adapt to the inconsistent gradients and Craft-ME cannot.
    \item 2. ME attack Performance (without training data only) of Vgg11 on 5\% subset of FEMNIST dataset (62-class), following the same setting as leaf benchmark \cite{caldas2018leaf}' online document (Table~\ref{tab:other_vgg11_femnist}). We observe a similar trend as in VGG-11 on CIFAR-10 experiments.
    \item 3. ME attack Performance (without training data only) of MobileNetV2 on CIFAR-10 (Table~\ref{tab:other_mob_cifar10}). We observe a similar trend as in VGG-11 on CIFAR-10 experiments.
\end{itemize}

\subsection{Time Cost Evaluation}

We evaluate time cost of five attacks on VGG-11 CIFAR-10 model (fine-tuning case).
The time cost measurement is done on a PC with a R7-5800X CPU and a single RTX-3090 GPU.

\begin{table}[htbp]
\caption{Time costs of proposed five attacks of attacking VGG-11 on CIFAR-10 (N=8) in fine-tuning case.}
\label{tab:time_requirement}
\begin{center}
\begin{small}
\resizebox{1.0\linewidth}{!}{
\begin{tabular}{lccccc} 
 \toprule
 Time Cost (s) & Craft &GAN &GM &Train &SoftTrain \\
 \midrule
  Preparation  &317.8	&44.5	&30.7	&4.7	&18.9\\
  Surrogate Training& 381.2	&339.3	&5523.1 &313.4	&949.5\\
  Total& 699.0	&383.8	&5553.8	&318.1	&968.4\\
 \bottomrule
\end{tabular}
}
\end{small}
\end{center}
\end{table}

\cref{tab:time_requirement} provides time cost breakdown for two phases, namely, preparation phase and training the surrogate model phase. The preparation phase includes crafting inputs in Craft-ME, fitting conditional GAN in GAN-ME, and crafting soft labels in SoftTrain-ME. From the results, we can see the Craft-ME needs the most preparation time and GAN-ME ranks the second. Both require generating crafted data and training the generator using collected gradients. For  training the surrogate model, GM-ME method requires the most time as solving the gradient matching involves  computation of second-order derivatives. Soft-Train method also spends more time compared to Craft-, GAN- and Train-ME because the soft-labels are used as the second objective.

In all the cases, the time cost of the proposed ME attacks is dominated by the cost of training the surrogate model. This heavily depends on the network topology, the number of iterations, and input size and vary from application to application, making it difficult to provide a comprehensive time complexity analysis.

\subsection{Conditional-GAN Architecture}
\label{sec:cgan}
\begin{figure*}[htbp]
\centering
\includegraphics[width=0.8\linewidth]{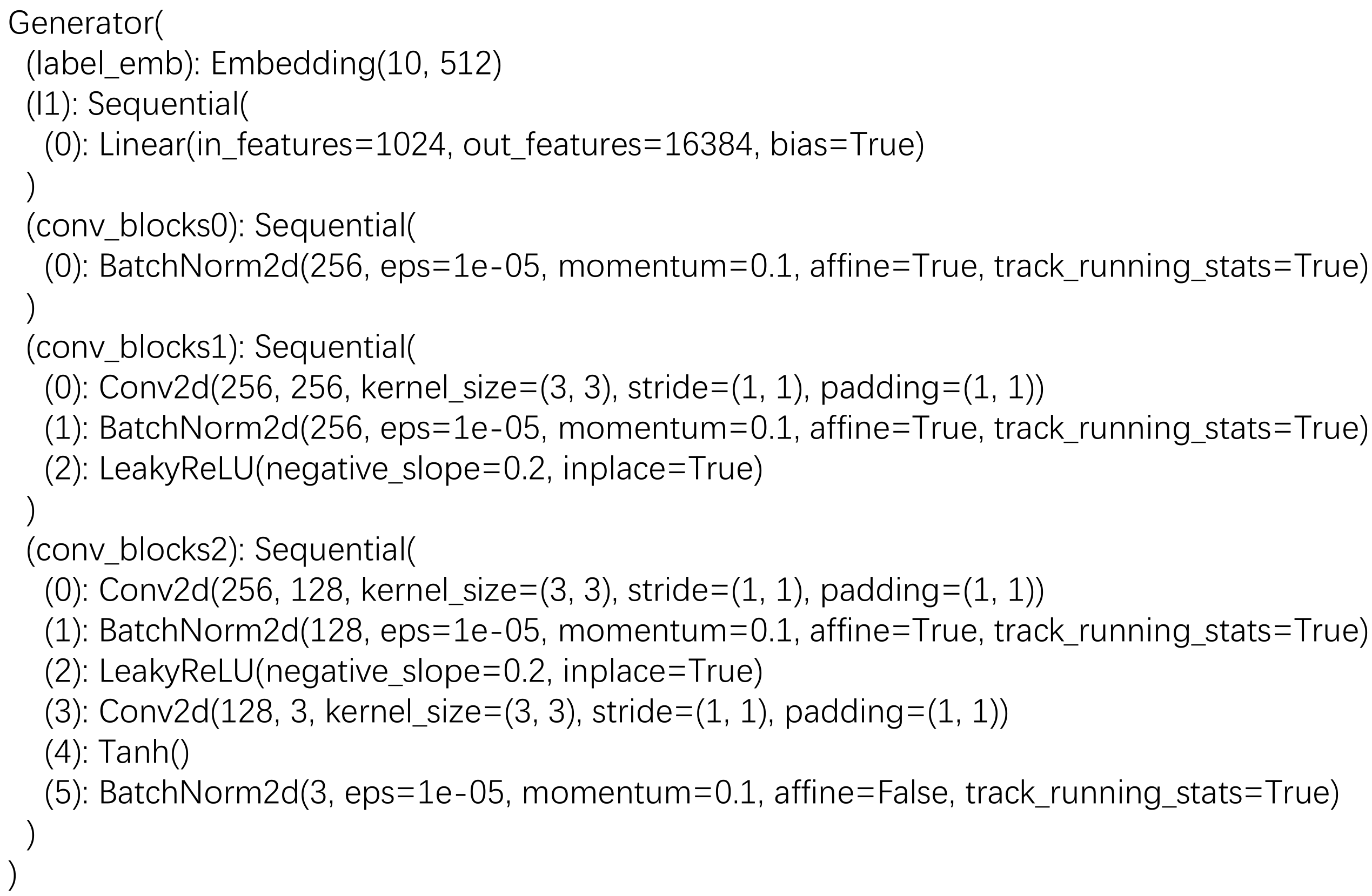}
\caption{Architecture detail of the c-GAN in GAN-ME.}
\vspace{-1.0em}
\label{fig:cgan_arch}
\end{figure*}

The detailed architecture of the conditional-GAN for GAN-ME attack is shown in Fig.~\ref{fig:cgan_arch}.

\newpage
\section{Model Inversion Attack Implementation}

\subsection{Model Inversion Attack Setting}
\label{apx:MI_attack}

\begin{figure*}[htbp]
\centering
\includegraphics[width=1.0\linewidth]{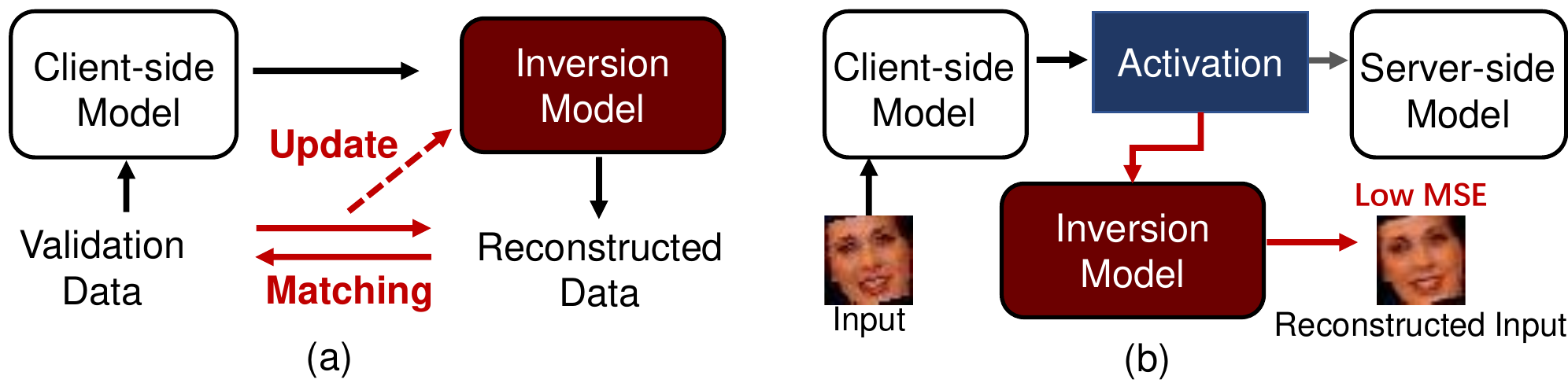}
\caption{Details of model inversion attack using L3 inversion model and the available validation dataset, done by an honest-but-curious server. (a) Train the inversion model on the validation dataset. (b) Use the inversion model to invert intermediate activation sent by clients.}
\vspace{-1.0em}
\label{fig:MI_attack}
\end{figure*}

The degree of how well client's data is protected in SFL is evaluated using Mean Squared Error (MSE) between ground-truth images and reconstructed images in Model Inversion Attack~(MIA). For MIA, we follow the same model-based attack methodology as in \cite{vepakomma2020nopeek, li2022ressfl}. 
The MIA flow is shown in Fig.~\ref{fig:MI_attack}. We assume the honest-but-curious attacker (this time, the server) has access to the 10K validation dataset of CIFAR-10. We use the L3 inversion model in \cite{li2022ressfl} to perform MIA, and use the trained L3 inversion model to reconstruct the raw image from the intermediate activation sent by benign clients.

\end{document}